\documentclass[10pt,unnumberedsections]{LTJournalArticle_mod}

\usepackage[switch]{lineno}

\usepackage{amsmath}
\usepackage{amssymb}
\usepackage{enumitem}
\usepackage[utf8]{inputenc}
\usepackage{hyperref}

\usepackage{dblfloatfix}

\usepackage[table]{xcolor}
\usepackage{graphicx}
\newcommand{\bx}{\boldsymbol{x}}

\newcommand{\bv}{\boldsymbol{v}}
\newcommand{\sv}{u}
\newcommand{\bsv}{\boldsymbol{\sv}}

\allowdisplaybreaks
\title{\LARGE The Clever Hans Effect in Unsupervised Learning}

\author{Jacob Kauffmann$^{1,2}$, Jonas Dippel$^{1,2,3}$, Lukas Ruff$^{1,3}$, Wojciech Samek$^{1,2,5}$,\\Klaus-Robert M\"uller$^{1,2,6,7,8,\star}$ and Gr\'egoire Montavon$^{4,2,1,\star}$}

\date{\small 
$^1$Department of Electrical Engineering and Computer Science, Technische Universit\"at Berlin, Germany\\
$^2$BIFOLD -- Berlin Institute for the Foundations of Learning and Data, Berlin, Germany\\
$^3$Aignostics, Berlin, Germany\\
$^4$Department of Mathematics and Computer Science, Freie Universit\"at Berlin, Germany\\
$^5$Department of Artificial Intelligence, Fraunhofer HHI, Berlin, Germany\\
$^6$Department of Artificial Intelligence, Korea University, Seoul, Korea\\
$^7$Max-Planck Institute for Informatics, Saarbr\"ucken, Germany\\
$^8$Google Deepmind, Berlin, Germany\\
$^\star$\,Corresponding authors (email: \texttt{klaus-robert.mueller@tu-berlin.de}, \texttt{gregoire.montavon@fu-berlin.de})
}

\renewcommand{\maketitlehookd}{%
\begin{abstract} 
Unsupervised learning has become an essential building block of AI systems. The representations it produces, e.g.\ in foundation models, are critical to a wide variety of downstream applications. It is therefore important to carefully examine unsupervised models to ensure not only that they produce accurate predictions, but also that these predictions are not ``right for the wrong reasons'', the so-called Clever Hans (CH) effect. Using specially developed Explainable AI techniques, we show for the first time that CH effects are widespread in unsupervised learning. Our empirical findings are enriched by theoretical insights, which interestingly point to inductive biases in the unsupervised learning machine as a primary source of CH effects. Overall, our work sheds light on unexplored risks associated with practical applications of unsupervised learning and suggests ways to make unsupervised learning more robust.
\end{abstract}
}

\begin{document}

\maketitle

Unsupervised learning is a subfield of machine learning that has gained prominence in recent years \cite{DBLP:conf/nips/BrownMRSKDNSSAA20, DBLP:journals/pieee/RuffKVMSKDM21, Krishnan2022}. It addresses fundamental limitations of supervised learning, such as the lack of labels in the data or the high cost of acquiring them. Unsupervised learning has achieved successes in modeling the unknown, such as uncovering new cancer subtypes \cite{Li2009, Jiang2019} or extracting novel insights from large historical corpora \cite{eberle-dh}.  Furthermore, the fact that unsupervised learning does not rely on task-specific labels makes it a good candidate for core AI infrastructure: Unsupervised anomaly detection provides the basis for various quality or integrity checks on the input data \cite{DBLP:books/sp/19/RettigKCP19,DBLP:books/sp/02/EskinAPPS02,DBLP:journals/ijcv/BergmannBFSS21,DBLP:journals/candie/ZipfelVFWKZ23}. Unsupervised learning is also a key technology behind `foundation models' \cite{Bommasani2021,DBLP:conf/nips/BrownMRSKDNSSAA20,simclr-v2,DBLP:conf/icml/RadfordKHRGASAM21,Moor2023,dippel2024foundation} which extract representations upon which various downstream models (e.g.\ classification, regression, `generative AI', etc.) can be built. The growing popularity of unsupervised learning models creates an urgent need to carefully examine how they arrive at their predictions. This is essential to ensure that potential flaws in the way these models process and represent the input data are not propagated to the many downstream supervised models that build upon them.

\begin{figure*}[t!]
    \centering
    \includegraphics[width=.95\textwidth]{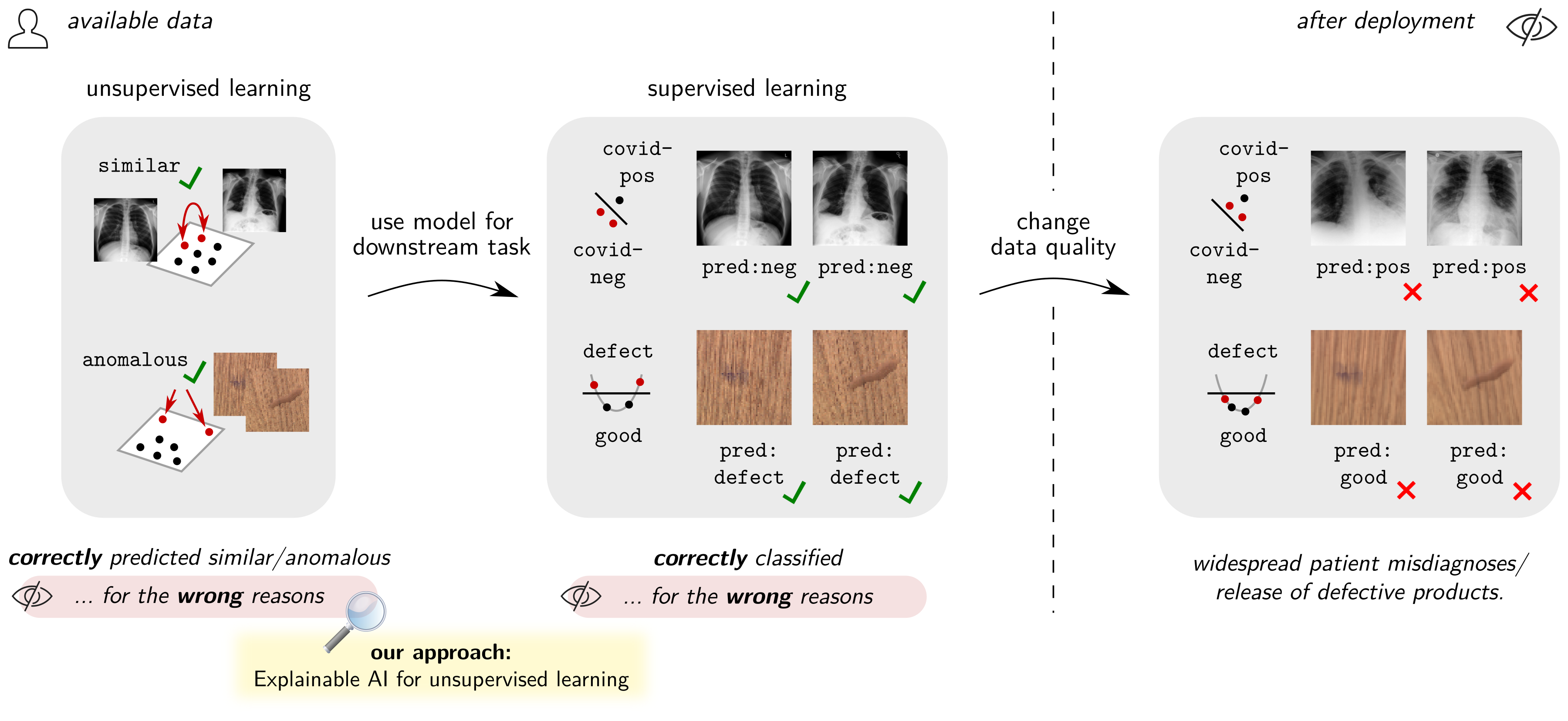}
    \caption{The CH effect in unsupervised learning. The unsupervised model correctly represents data instances as similar or anomalous, but for the wrong reasons, e.g.\ based on irrelevant input features such as artifacts. The problem is critical because the flaw can be inherited by potentially many downstream tasks. The CH effect typically goes undetected in a classical validation procedure and manifests itself in the form of prediction errors only after deployment.}
    \label{fig:critical}
\end{figure*}

In this paper, we show for the first time that unsupervised learning models largely suffer from Clever Hans (CH) effects \cite{Lapuschkin2019}, also known as ``right for the wrong reasons''. Specifically, we find that unsupervised learning models often produce representations from which instances can be correctly predicted to be e.g.\ similar or anomalous, although largely supported by data quality artifacts. The flawed prediction strategy is not detectable by common evaluation benchmarks such as cross-validation, but may manifest itself much later in `downstream' applications in the form of unexpected errors, e.g.\ if subtle changes in the input data occur after deployment (cf.\ Fig.\ \ref{fig:critical}). While CH effects have been studied quite extensively for \textit{supervised learning} \cite{Lapuschkin2019, geirhos2020shortcut, DBLP:journals/inffus/AndersWNSML22, linhardt2023preemptively, Winkler2019}, the lack of similar studies in the context of \textit{unsupervised learning}, together with the fact that unsupervised models supply many downstream applications, is a cause for concern.

For example, in industrial inspection, which often relies on unsupervised anomaly detection \cite{DBLP:journals/ijcv/BergmannBFSS21, DBLP:journals/candie/ZipfelVFWKZ23}, we find that a CH decision strategy can systematically miss a wide range of manufacturing defects, resulting in potentially significant costs. As another example, unsupervised foundation models, advocated in the medical domain to provide robust features for various specialized diagnostic tasks, can potentially introduce CH effects into many of these tasks, with the prominent risk of large-scale misdiagnosis. These scenarios (illustrated in Fig.\ \ref{fig:critical}) highlight the practical implications of an unsupervised CH effect, which, unlike its supervised counterpart, may not be limited to malfunctioning in a single specific task, but potentially in all downstream tasks.

To uncover and understand unsupervised Clever Hans effects, we propose to use Explainable AI \cite{montavon2018methods,gunning2019xai, arrieta2020xai, samek2021explaining,klauschen2023toward} (here techniques that build on the LRP explanation framework \cite{bach-plos15, DBLP:series/lncs/MontavonBLSM19, DBLP:journals/pr/KauffmannMM20}). Our proposed use of these techniques allows us to identify at scale which input features are used (or misused) by the unsupervised ML model, without having to formulate specific hypotheses or downstream tasks. Specifically, we use an extension of LRP called BiLRP \cite{eberle-dh} to reveal input patterns that are jointly responsible for similarity in the representation space. We also combine LRP with `virtual layers' \cite{vielhaben23,chormai24} to reveal pixel and frequency components that are jointly responsible for predicted anomalies.

Furthermore, our Explainable AI-based analysis allows us to pinpoint more formal causes for the emergence of unsupervised CH effects. In particular, they are due not so much to the \textit{data}, but to the \textit{unsupervised learning machine}, which hinders the integration of the true task-supporting features into the model, even though vast amounts of data points are available. Our findings provide a novel direction for developing targeted strategies to mitigate CH effects and increase model robustness.

Overall, our work sheds light on the presence, prominence, and distinctiveness of CH effects in unsupervised learning, calling for increased scrutiny of this essential component of modern AI systems.

\section{Results}

Using Explainable AI, we investigate the emergence of Clever Hans effects in a representative set of unsupervised learning tasks, including representation learning and anomaly detection.

\subsection{Clever Hans Effects in Representation Learning}
\label{section:results-xray}

\addtocounter{figure}{1}
\begin{figure*}[b!]
    \centering
    \includegraphics[width=\textwidth]{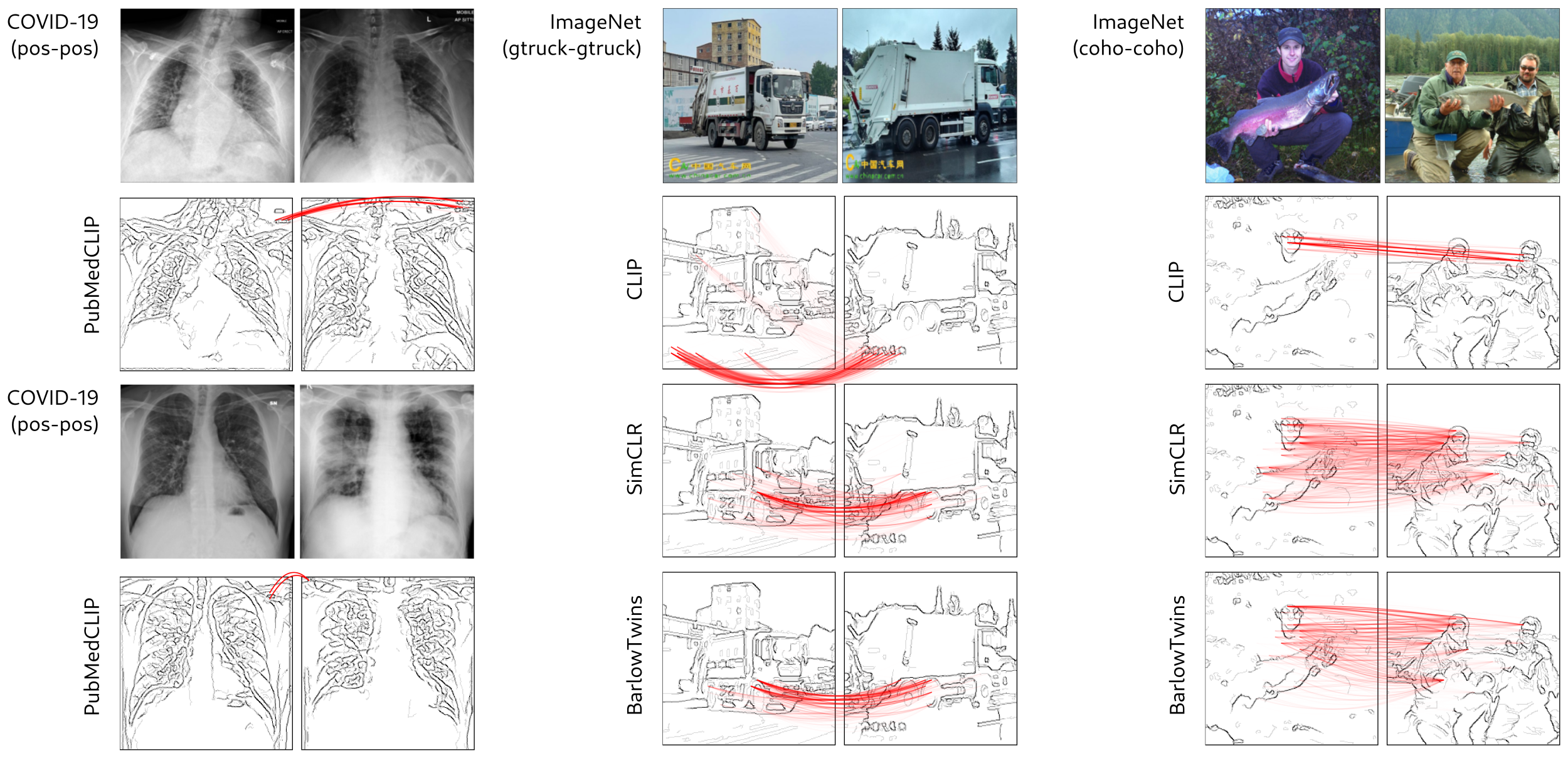}
     \caption{BiLRP analysis of the PubMedCLIP unsupervised model and the general-purpose CLIP, SimCLR, and BarlowTwins unsupervised models. The analysis reveals a variety of CH strategies, specifically instances that are correctly represented as similar by the unsupervised model but for the wrong reasons. Similarity between COVID-positive instances arises from shared spurious textual annotations. Similarly, on ImageNet data, unsupervised models incorrectly rely on logo artifacts or the presence of humans in the background for their similarity predictions.}
     \label{fig:bilrp-results}
\end{figure*}
\addtocounter{figure}{-2}

We first consider an application of representation learning in the context of detecting \mbox{COVID-19} cases from X-ray scans \cite{Wang2020,DeGrave2021}. Simulating an early pandemic phase characterized by data scarcity, similar to \cite{DeGrave2021}, we use a dataset aggregation approach where a large, well-established non-COVID-19 dataset is merged with a more recent COVID-19 dataset aggregated from multiple sources. Specifically, we aggregate 2597 instances of the CXR8 dataset \cite{Wang2017} from the National Institute of Health (NIH), collected between 1992 and 2015, with the 535 instances of the GitHub-hosted `COVID-19 image data collection' \cite{cohen2020covidProspective}. We refer to these subsets as `NIH' and `GitHub', respectively.

Further motivated by the need to accommodate the critically small number of COVID-19 instances in the aggregated dataset and to avoid overfitting, we choose to rely on the representations provided by unsupervised foundation models \cite{DBLP:conf/iccv/AziziMRBFDLKKCN21,DBLP:conf/eacl/EslamiMM23,Huang2023,dippel2024foundation}. Specifically, we feed our data into a pre-trained PubMedCLIP model \cite{DBLP:conf/eacl/EslamiMM23}, which has built its representation from a very large collection of X-ray scans in an unsupervised manner. Based on the PubMedCLIP representation, we train a downstream classifier that separates COVID-19 from non-COVID-19 instances with a class-balanced accuracy of 88.6\% on the test set (cf.\ Table \ref{table:robustness}). However, a closer look at the structure of this performance score reveals a strong disparity between the NIH and GitHub subgroups, with all NIH instances being correctly classified and the GitHub instances having a lower class-balanced accuracy of 75.2\%, and, more strikingly, a false positive rate (FPR) of 40\%, as shown in Table \ref{table:robustness}. Considering that the higher heterogeneity of instances in the GitHub dataset is more characteristic of real-world conditions, such a high FPR prohibits any practical use of the model as a diagnostic tool. We emphasize that this deficiency could have been easily overlooked if one did not pay close attention to (or did not know) the data sources, and instead relied only on the overall accuracy score.

\begin{figure}[t!]
    \centering
    \includegraphics[width=\linewidth]{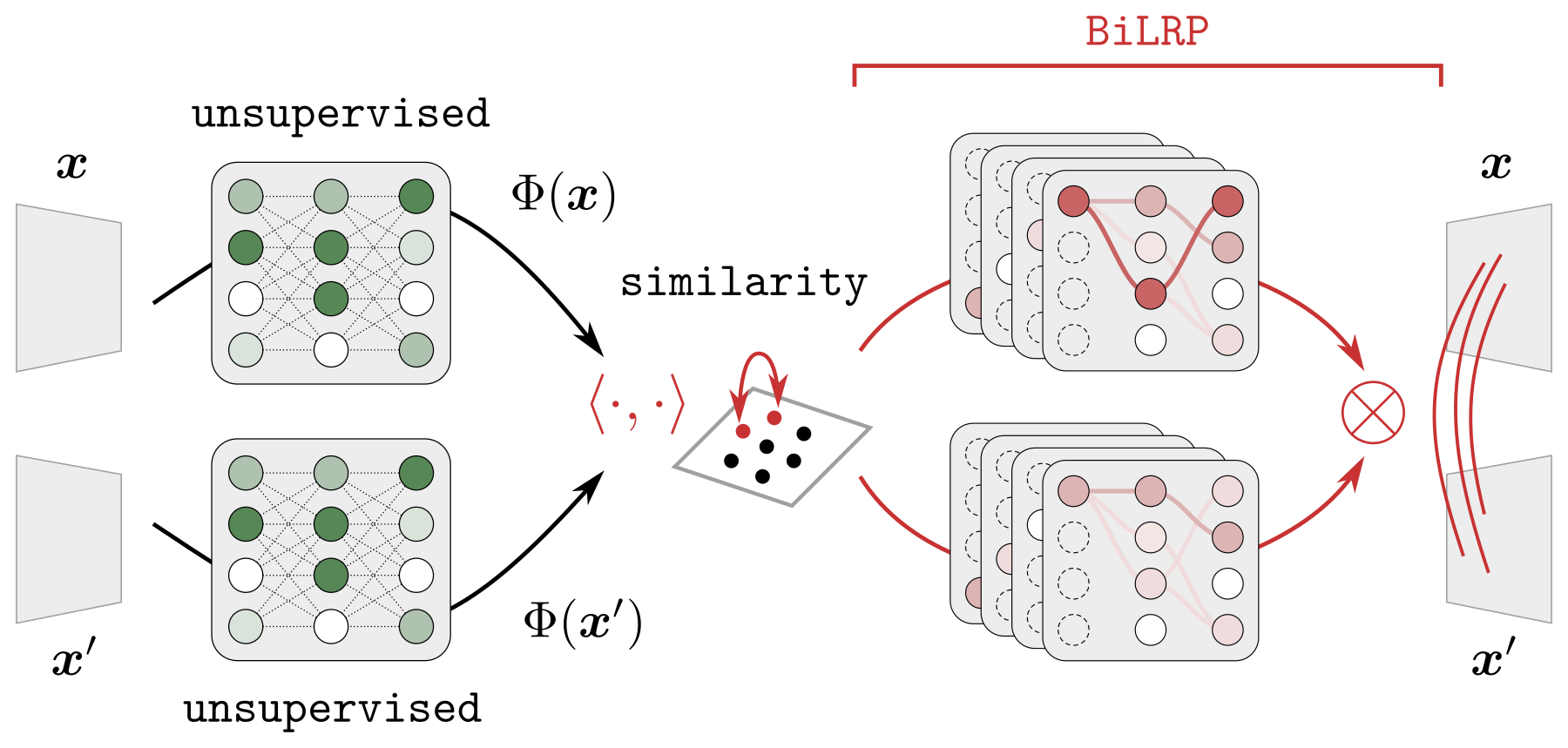}
    \caption{Illustration of the BiLRP method for explaining similarity predictions of a representation learning model. The output of BiLRP is a decomposition of the predicted similarity onto pairs of features from the two input images. It is typically displayed as a weighted bipartite graph connecting the contributing feature pairs.}
    \label{fig:bilrp-method}
\end{figure}
\addtocounter{figure}{1}

To proactively detect this heterogeneous, non-robust prediction behavior and to test whether it originates from the unsupervised PubMedCLIP component, we propose to use Explainable AI. Using the BiLRP  \cite{eberle-dh} technique, we investigate whether PubMedCLIP represents different instances as similar based on a correct strategy or a `Clever Hans' strategy. The BiLRP method is illustrated in Fig.\ \ref{fig:bilrp-method} and its mathematical formulation is given in the Methods section. The output of BiLRP for two exemplary pairs of COVID-positive instances is shown in Fig.\ \ref{fig:bilrp-results} (left). The BiLRP explanation shows that the modeled similarity between pairs of COVID-positive instances comes from text-like annotations that appear in both images. This is a clear case of a CH effect, where the modeled similarity is  \textit{right} (instances share being COVID-positive), but for the \textit{wrong reasons} (similarity is based on shared textual annotations). The unmasking of this CH effect by BiLRP warns of the risk that downstream models based on these similarities (and more generally on the flawed representation) may inherit the CH effect. This a posteriori explains the excessively high (40\%) FPR of the downstream classifier on the Github subset, as the model's reliance on text makes it difficult to separate Github's COVID-positive from COVID-negative instances (cf.\ Supplementary Note A for further analysis). We note that, unlike the per-group accuracy analysis above, our BiLRP analysis did not require provenance metadata (GitHub or NIH), nor did it focus on a specific downstream task with its specific labels.

To test whether representation learning has a general tendency to evolve CH strategies beyond the above use case, we downloaded three generic foundation models, namely the original CLIP model \cite{DBLP:conf/icml/RadfordKHRGASAM21}, SimCLR \cite{simclr-v1,simclr-v2} and BarlowTwins \cite{DBLP:conf/icml/ZbontarJMLD21}. As a downstream task, we consider the classification, using linear-softmax classifiers, of the 8 classes from ImageNet \cite{DBLP:conf/cvpr/DengDSLL009} that share the WordNet ID `truck' and of the 16 ImageNet classes that share the WordNet ID `fish' (see the Methods section for details). The test accuracy of each model on these two tasks is given in Table \ref{table:robustness} (columns `original'). On the truck classification task, the CLIP model performs best, with an accuracy of 85.0\%. On the fish classification task, the CLIP and supervised models perform best, with accuracies of 86.5\% and 86.2\%, respectively.

We use BiLRP to examine the representations of these unsupervised models. In Fig.\ \ref{fig:bilrp-results} (center), we observe that CLIP-based similarities, as in PubMedCLIP, also rely on text. Here, a textual logo in the lower left corner of two garbage truck images is used to support similarity\footnote{The reliance on logos for the garbage truck class has been highlighted in previous work in the context of supervised learning  \cite{DBLP:journals/inffus/AndersWNSML22}.}. In contrast, SimCLR and BarlowTwins instead rely on the actual garbage truck, demonstrating that CH effects are specific to the unsupervised learning technique. In Fig.\ \ref{fig:bilrp-results} (right), we observe that \textit{all} unsupervised models focus on humans and are unresponsive to fish features. CLIP focuses specifically on facial features. While the detection of human features may be useful for many downstream tasks, the suppression of fish features strongly exposes other downstream fish classification tasks to even mild spurious correlations, systematically leading to CH behavior.

The heterogeneity of strategies across models revealed by BiLRP suggests that, in addition to inhomogeneities in the unsupervised data, biases induced by the unsupervised pre-training algorithm\footnote{See also \cite{chen-feature-sup, robinson-shortcuts, dippel2021towards, li2023addressing} for further analyses of these biases.} play an important role in the formation of CH effects. The spurious amplification of humans in the center of the image observed for SimCLR and BarlowTwins can be attributed to the cropping mechanism these methods implement, where human features in the center are sufficient and robust to solve the SimCLR and BarlowTwins similarity tasks. In the case of CLIP (and PubMedCLIP), the systematic amplification of textual, facial, or other identifying features can be explained by the fact that these features often provide useful information for CLIP's unsupervised image-text matching task.

The consequences of these Clever Hans effects are shown \textit{quantitatively} in Table \ref{table:robustness}. We observe a systematic degradation of performance when moving from the original data to data subsets where the presence/absence of an artifact is no longer predictive of the class. In particular, for the truck data, we observe a sharp drop in the accuracy of the CLIP model from 85.0\% on the original data to 80.5\% on the same data with a logo inserted on each image (column `logo' in Table \ref{table:robustness}). For the fish data, a similar drop in accuracy is observed for SimCLR and BarlowTwins from 82.2\% and 83.1\% respectively on the original data to 78.6\%  and 75.6\% respectively when only images containing humans are retained and class rebalancing is performed. This shows that the CH effects detected by our BiLRP analysis have concrete negative practical consequences in terms of the ability of unsupervised models and downstream classifiers to predict uniformly well across subgroups and to generalize. In comparison, the baseline supervised model generally shows more stable performance between the original data and the newly defined subsets. A detailed analysis of the structure of the prediction errors for each classification task, supported by confusion matrices, is given in Supplementary Note B.

\addtocounter{figure}{1}
\begin{figure*}[b!]
    \centering
\includegraphics[page=1]{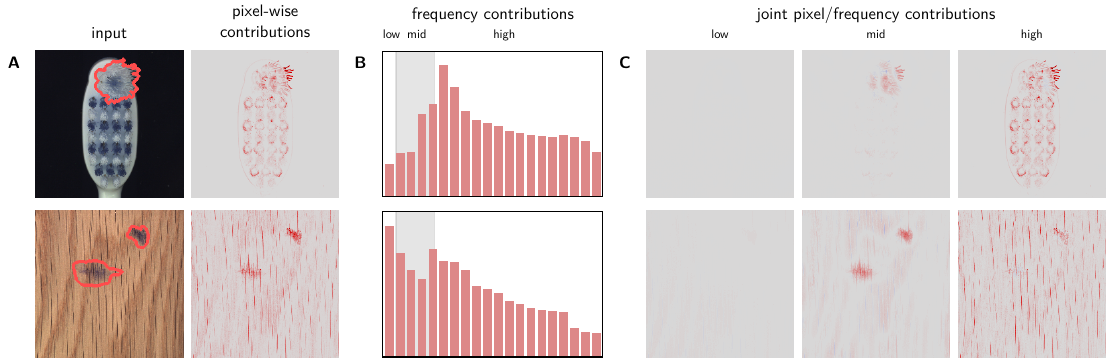}
    \caption{Explainable AI analysis of the D2Neighbors anomaly model. {\bf A:} Selected images from MVTec-AD (ground truth anomaly shown as red contour) and a pixel-wise LRP explanation of the anomaly prediction. Explanations for other categories are given in Supplementary Note C. {\bf B:} Frequency domain explanations (averaged over all anomalous test instances). The x-axis represents the frequencies (on a power scale), and the y-axis is the contribution of the corresponding frequencies to the anomaly prediction. {\bf C:} Pixel-wise contributions filtered by frequency band.}
    \label{fig:joint}
\end{figure*}
\addtocounter{figure}{-2}

\subsection{Clever Hans Effects in Anomaly Detection}
\label{section:results-mvtec}

Extending our investigation of the CH effect to another instance of unsupervised learning, namely anomaly detection, we consider an industrial inspection use case based on the popular MVTec-AD dataset \cite{DBLP:journals/ijcv/BergmannBFSS21}. The dataset consists of 15 product categories, each consisting of a training set of images without manufacturing defects and a test set of images with and without defects. Since manufacturing defects are infrequent and heterogeneous in nature, the problem is typically approached using unsupervised anomaly detection \cite{DBLP:journals/ijcv/BergmannBFSS21,DBLP:journals/pieee/RuffKVMSKDM21}. These models map each instance to an anomaly score, from which threshold-based downstream models can be built to classify between instances with and without manufacturing defects. Unsupervised anomaly detection has received considerable attention, with sophisticated approaches based on deep neural networks such as PatchCore \cite{DBLP:conf/cvpr/RothPZSBG22} or EfficientAD \cite{DBLP:conf/wacv/BatznerHK24}, showing excellent performance in detecting a wide range of industrial defects.

Somewhat surprisingly, simpler approaches based on distances in pixel space show comparatively high performance for selected tasks \cite{DBLP:journals/pieee/RuffKVMSKDM21}. We consider one such approach, which we call `D2Neighbors', where anomalies are predicted according to the distance to neighbors in the training data. Specifically, the anomaly score of an instance $\bx$ is computed as $f(\bx) = \text{softmin}_j\{ \|\bx - \bsv_j\|^2\}$ where $(\bsv_j)_{i=1}^N$ is the set of available inlier instances (see the Methods section for details on the model and data preprocessing). This anomaly model belongs to the broader class of distance-based models \cite{HARMELING20061608,DBLP:books/sp/Aggarwal2013,DBLP:journals/csur/ChandolaBK09}, and connections can be made to kernel density estimation \cite{Parzen1962,DBLP:journals/jmlr/KimS12} and one-class SVMs \cite{DBLP:journals/neco/ScholkopfPSSW01}. Using D2Neighbors, we are able to build downstream models that classify industrial defects of the MVTec data with F1-scores above 0.9 for five categories (bottle, capsule, pill, toothbrush, and wood).

\begin{figure}[t!]
    \centering
\includegraphics[page=3]{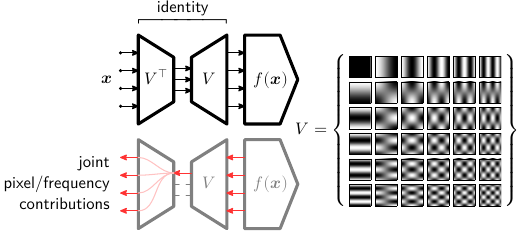}
    \caption{Schematic of the virtual inspection layer used to explain anomalies in the joint pixel-frequency domain. To map pixels to frequencies and back, we use the discrete cosine transform (DCT), whose basis elements are shown on the right.}
    \label{fig:vlayer}
\end{figure}
\addtocounter{figure}{1}

\begin{table*}[b!]
\small \centering
\newcommand{\pdecrease}[1]{$\phantom{\downarrow}$\,#1\,$\color{red}\downarrow$}
\newcommand{\pincrease}[1]{$\phantom{\uparrow}$\,#1\,$\color{green!50!black}\uparrow$}
\renewcommand{\arraystretch}{1.15}
\begin{tabular}{lcccccc}
\multicolumn{1}{l}{}
& \multicolumn{2}{c}{COVID-19} & \multicolumn{2}{c}{ImageNet:truck} & \multicolumn{2}{c}{ImageNet:fish}\\\midrule
& \textit{original} & github & \!\!\textit{original}\!\! & logo & \!\!\textit{original}\!\! & human\\\midrule
PubMedCLIP &  \textit{88.6}   & \pdecrease{75.2} & --- &--- &--- &---\\[-1mm]
 &  \scriptsize \!\!FPR: 14\%\!\!   & \scriptsize  \!\!FPR: 40\%\!\! &  &  & &\\
CLIP & --- &--- &  \textit{85.0}    & \pdecrease{80.5} & \textit{86.5}       & 83.8\\
\rowcolor{gray!20!white}~+~CH mitigation\!\!\!\! & --- &--- & \textit{85.2} & \pincrease{85.0} & --- & ---\\
SimCLR  & --- &---  & \textit{74.8}  & 74.5  & \textit{82.2}   & \pdecrease{78.6}  \\
BarlowTwins & --- &--- & \textit{80.2}  & 80.2  & \textit{83.1}    & \pdecrease{75.6} \\
\textit{Supervised}   & --- &--- & \textit{83.8}  &  83.2     & \textit{86.2}  &  84.2  \\\bottomrule
\end{tabular}%
~~~~~%
\begin{tabular}{lcc}
& \multicolumn{2}{c}{MVTec-AD$^\star$}\\\midrule
& \textit{original} & deployed\\\midrule
D2Neighbors ($\ell_2$) & \textit{0.92} & \pdecrease{0.80} \\[-1mm]
 & \scriptsize FNR: 4\% & \scriptsize FNR: 26\% \\
\rowcolor{gray!20!white}~+~CH mitigation & \textit{0.94} & \pincrease{0.93} \\
PatchCore &\textit{0.92} & \pdecrease{0.85} \\
\rowcolor{gray!20!white}~+~CH mitigation & \textit{0.97} & \pincrease{0.96} \\
D2Neighbors ($\ell_1$) & \textit{0.93} & \pdecrease{0.74} \\
D2Neighbors ($\ell_4$) & \textit{0.91} & \pdecrease{0.83}\\
\bottomrule
\end{tabular}

\caption{Performance of unsupervised models on different data subsets or conditions. \textit{Left:} Class-balanced classification accuracy of downstream models (COVID-19, truck and fish classification) built on different unsupervised representation learning models. Accuracy scores are reported for the original test set and for a subset containing specific instances. \textit{Right:} F1 scores of different anomaly detection models. Performance is reported on the simulated original and post-deployment data conditions (standard and antialiased resizing, respectively). $(^\star)$ Performance on MVTec-AD is computed for each of the 5 classes retained for the analysis and averaged. $({\color{red}\downarrow})$ Severe performance degradation (of 3 percentage points or more) after deployment. $({\color{green!50!black}\uparrow})$ Major performance improvement (of 3 percentage points or more) after CH mitigation.}
\label{table:robustness}
\end{table*}

To shed light on the prediction strategy that leads to these unexpectedly high F1-scores, we investigate the underlying D2Neighbors anomaly model using Explainable AI. Specifically, we consider an extension of LRP for anomaly detection \cite{DBLP:journals/pr/KauffmannMM20,Montavon2022} and further equip the explanation technique with `virtual layers' \cite{vielhaben23,chormai24}. The technique of `virtual layers' (cf.\ Fig.\ \ref{fig:vlayer}) is to map the input to an abstract domain and back, leaving the prediction function \emph{unchanged}, but providing a new representation in terms of which the prediction can be explained. We construct such a layer by applying the discrete cosine transform (DCT) \cite{10.3389/fnbot.2021.767299}, shown in Fig.\ \ref{fig:vlayer} (right), followed by its inverse. This allows us to explain the predictions \emph{jointly} in terms of pixels and frequencies.

The result of our proposed analysis is shown in Fig.\ \ref{fig:joint} for instances from two of the five retained MVTec categories (see Supplementary Note C for more instances). Explanations at the \textit{pixel} level show that D2Neighbors supports its anomaly predictions largely based on pixels containing the actual industrial defect. The squared difference in its distance function ($\|\Delta\|^2 = \sum_i \Delta_i^2$) encourages a sparse pixel-wise response of the model, efficiently discarding regions of the image where the new instance shows no difference from instances in the training data. However, we also see in the pixel-wise explanation that a non-negligible part of the anomaly prediction comes from irrelevant background pixels. Joint \textit{pixel-frequency} explanations shed light on these unresolved contributions, showing that they arise mostly from the high-frequency part of the model's decision strategy (Fig.\ \ref{fig:joint} B and C).

The exposure of the model to these irrelevant high-frequency features, as detected by our LRP analysis, raises concerns about the robustness of the model under changing data conditions (e.g.\ after the model is deployed). We experiment with an innocuous perturbation of the data conditions, which consists of changing the image resizing algorithm from OpenCV's nearest neighbor resizing to a more sophisticated resizing method that includes antialiasing. Resizing techniques have been shown in some cases to significantly affect image quality and ML model prediction \cite{DBLP:conf/cvpr/Parmar0Z22}, but their effect on ML models, especially unsupervised ones, has been little studied. The performance of the D2Neighbors model and other models before and after changing the resizing algorithm is shown in Table \ref{table:robustness} (columns `original' and `deployed' respectively). The F1 score performance of D2Neighbors degrades significantly from $0.92$ to $0.80$. In particular, there is a large increase in the false negative rate (FNR) from $4\%$ to $26\%$ (cf.\ Table \ref{table:robustness}). In an industrial inspection setting, an increase in FNR has serious consequences, in particular, many defective instances are missed and propagated through the production chain, resulting in wasted resources in subsequent production stages and high recall costs. This performance degradation of D2Neighbors in post-deployment conditions is particularly surprising given that data quality has actually \textit{improved}. This is a direct consequence of the CH strategy we identified, namely D2Neighbors' reliance on high frequencies. When antialiasing is introduced into the resizing procedure, the high frequencies that the D2Neighbors model uses to support its prediction disappear from the data, and this significantly reduces the anomaly scores of each instance, thereby increasing the FNR.

Interestingly, the performance degradation of D2Neighbors is even more severe when the $\ell_2$-norm in the distance function is replaced by the $\ell_1$-norm (cf.\ Table \ref{table:robustness}). This can be explained by the fact that the $\ell_1$-norm, unlike the $\ell_2$-norm, does not implement pixel-wise sparsity and is therefore more exposed to irrelevant input regions. Conversely, the application of an $\ell_4$-norm results in sensibly higher robustness than the original $\ell_2$-norm. We note that even the more sophisticated PatchCore model \cite{DBLP:conf/cvpr/RothPZSBG22} (cf.\ Methods section), which integrates several robustness mechanisms including preprocessing layers and spatial max-pooling, is not fully immune to irrelevant high-frequency components and also shows a noticeable performance degradation in post-deployment conditions (cf.\ Table \ref{table:robustness}).

Overall, our experiments show that performance on the available test data is an insufficient indicator of the quality of an anomaly detector. Its robustness to changes in the quality of the input data cannot be guaranteed. This can be diagnosed by Explainable AI and a careful subsequent inspection of the structure of the anomaly model.

\subsection{Alleviating CH in Unsupervised Learning}

Leveraging the Explainable AI analysis above, we aim to build models that are more robust across different data subgroups and in post-deployment conditions. Unlike previously proposed CH removal techniques \cite{DBLP:journals/inffus/AndersWNSML22, linhardt2023preemptively}, we aim to operate on the unsupervised model rather than the downstream tasks. This potentially allows us to achieve broad robustness improvements while leaving the downstream learning machines (training supervised classifiers or adjusting detection thresholds) untouched.

We first test our CH mitigation approach on the CLIP model, which our Explainable AI analysis has shown to incorrectly rely on text. Specifically, we focus on an early layer of the CLIP model (encoder.relu3) and remove the feature maps that are most responsive to text. We then apply a similar CH mitigation approach to anomaly detection, which our Explainable AI analysis has shown to be overexposed to high frequencies. Here, we propose to prune the high frequencies by inserting a blur layer at the input of the model.

In both cases, the proposed CH mitigation technique provides strong benefits in terms of model robustness. As shown in Table \ref{table:robustness}, rows `CH mitigation', our robustified models substantially reverse the performance degradation observed in post-deployment conditions, reaching performance levels close to, and in some cases superior to, those measured on the original data.

\section{Discussion}

Unsupervised learning is an essential category of ML that is increasingly used in core AI infrastructure to power a variety of downstream tasks, including generative AI. Much research to date has focused on improving the performance of unsupervised learning algorithms, for example, to maximize accuracy scores in downstream classification tasks. These evaluations often pay little attention to the exact strategy used by the unsupervised model to achieve the reported high performance, in particular whether these models rely on Clever Hans strategies.

Building on recent techniques from Explainable AI, we have shown for the first time that CH strategies are prevalent in several unsupervised learning paradigms. These strategies can take various forms, such as correctly predicting the similarity of two X-ray scans based on irrelevant shared annotations, or predicting that an image is anomalous based on small but widespread pixel-level artifacts. These flawed prediction strategies result in models that do not transfer well to changing conditions at test time. Most importantly, their lack of robustness infects many downstream models that rely on them. As we have shown  in two use cases, this can lead to widespread misdiagnosis of patients or failure to detect manufacturing defects. 

Therefore, addressing CH effects is a critical step towards reliable use of unsupervised learning methods. However, compared to a purely task-specific supervised approach, the addition of an unsupervised component, potentially serving multiple downstream tasks, adds another dimension of complexity to the modeling problem. In particular, one must decide whether to handle CH effects in the downstream classifier or \textit{directly} in the unsupervised model part. Approaches consisting of dynamically updating downstream models in response to changing conditions \cite{sugiyama2007covariate,sugiyama2012machine,DBLP:conf/nips/IwasawaM21,Esposito2021}, or revising their decision strategies with human feedback \cite{DBLP:journals/inffus/AndersWNSML22,DBLP:conf/iclr/KirichenkoIW23,linhardt2023preemptively} are possible solutions to maintain high accuracy. However, these approaches may not be sustainable because CH mitigation must be performed repeatedly for each new downstream task. The problem may persist even after a flaw in the foundation model becomes known (e.g. \cite{DBLP:conf/acl/NivenK19,Heinzerling2020NLPsCH}), due to release intervals and the high cost of retraining these models. Instead, our results argue for addressing CH effects directly when building the unsupervised model, with the goal of achieving persistent robustness that benefits existing and future downstream applications.

Using advanced Explainable AI techniques targeted at unsupervised learning, we have uncovered multiple and diverse CH effects, such as the reliance of CLIP (and its derivative PubMedCLIP) on textual annotations, or a systemic over-exposure of unsupervised anomaly models to noise. Interestingly, our analysis has revealed that these unsupervised CH effects differ from supervised ones in that they arise less from the data and more from inductive biases in the model and learning algorithm. These include spurious suppression/amplification effects caused by the representation learning's training objective, or a failure of unsupervised anomaly detection architectures to replicate frequency filtering mechanisms found in supervised learning \cite{DBLP:journals/jmlr/BraunBM08, DBLP:conf/icml/BasriGGJKK20, DBLP:conf/nips/Fridovich-KeilL22}, leaving the learned models highly exposed to noise.  Furthermore, our Explainable AI analysis not only provided new insights into the formal causes of unstable behavior in unsupervised learning. Our experiments also showed how pruning the high-frequency or spuriously amplified features revealed by our Explainable AI analysis leads to systematic performance improvements on difficult data subgroups or in post-deployment conditions. In doing so, we have demonstrated the actionability of our Explainable AI approach, showing that it can guide the process of identifying and subsequently correcting the faulty components of an unsupervised learning model.

While our investigation of unsupervised CH effects and their consequences has focused on image data, extension to other data modalities seems straightforward. Explainable AI techniques such as LRP, which are capable of accurate dataset-wide explanations, operate independently of the type of input data. They have recently been extended to recurrent neural networks \cite{DBLP:series/lncs/ArrasAWMGMHS19}, graph neural networks  \cite{schnake2022higher}, transformers \cite{DBLP:conf/icml/AliSEMMW22}, and state space models \cite{jafari2024mambalrp}, which represent the state of the art for large language models and other models of structured data. Thus, our analysis could be extended to analyze other instances of unsupervised learning, such as anomaly detection in time series or the representations learned by large language models (e.g.\ \cite{DBLP:journals/access/MunirSDA19, DBLP:conf/naacl/DevlinCLT19}).

Overall, we believe that the CH effect in unsupervised learning, and the uncontrolled risks associated with it, is a question of general importance, and that Explainable AI and its recent developments provide an effective way to tackle it.

\section{Methods}

This section first introduces the unsupervised ML models studied in this work and the datasets on which they are applied. It then presents the layer-wise relevance propagation (LRP) method for explaining predictions, its BiLRP extension for explaining similarity, and the technique of `virtual layers' for generating joint pixel-frequency explanations.

\subsection{ML Models and Data for Representation Learning} 

Representation learning experiments were performed on the SimCLR \cite{simclr-v1, simclr-v2}, CLIP\cite{DBLP:conf/icml/RadfordKHRGASAM21}, BarlowTwins \cite{DBLP:conf/icml/ZbontarJMLD21} and PubMedCLIP \cite{DBLP:conf/eacl/EslamiMM23} models. All include a version based on the ResNet50 architecture \cite{he-resnet} and come with pre-trained weights. SimCLR augments the input images with resized crops, color jitter, and Gaussian blur to create two different views of the same image. These views are then used to create positive and negative pairs, where the positive pairs represent the same image from two different perspectives, and the negative pairs are created by pairing different images. The contrastive loss objective maximizes the similarity between the representations of the positive pairs while minimizing the similarity between the representations of the negative pairs. Barlow Twins is similar to SimCLR in that it also generates augmented views of the input image through random resized crops and color augmentation, and maximizes their cosine similarity in representation space. However, it differs from SimCLR in the exact mechanisms used to prevent representation collapse.  In our experiments, we use the weights from the vissl\footnote{\url{https://vissl.ai/}} library. CLIP (and its derivative PubMedCLIP) learns representations by using a large collection of image-text pairs from the Internet. Images are given to an image encoder, and the corresponding texts are given to a text encoder. The similarity of the two resulting embeddings is then maximized with a contrastive loss. In our experiments, we use the ResNet-50 weights from OpenAI\footnote{\url{https://github.com/openai/CLIP}} for CLIP. PubMedCLIP\footnote{\url{https://github.com/sarahESL/PubMedCLIP}} is based on a pre-trained CLIP model fine-tuned on the ROCO dataset \cite{roco-dataset}, a collection of radiology and image caption pairs. For all representation learning experiments, the \textit{supervised baselines} share the same architecture as their unsupervised counterparts, but are trained in a purely supervised fashion using backpropagation.

The analysis and training of these models were performed on different datasets. The ImageNet experiments were performed on two ImageNet subsets. First, the `truck' subset consisting of the 8 classes sharing the WordNetID `truck' (minivan, moving van, police van, fire engine, garbage truck, pickup, tow truck and trailer truck). Then the `fish' subset consisting of the 16 classes sharing the WordNetID `fish' (tench, barracouta, coho, sturgeon, gar, stingray, great white shark, hammerhead, tiger shark, puffer, electric ray, goldfish, eel, anemone fish, rock beauty and lionfish). For the X-ray experiments, we combined the NIH ChestX-ray8 (CXR8) dataset\footnote{\url{https://academictorrents.com/details/e615d3aebce373f1dc8bd9d11064da55bdadede0}} and the GitHub-hosted `COVID-19 image data collection' \footnote{\url{https://github.com/ieee8023/covid-chestxray-dataset}}. The NIH dataset contributed 2597 negative images, while the GitHub dataset contained 342 COVID-positive and 193 negative images. All images were resized to 224x224 pixels and center-cropped. We split the patients 80:20 into training and test sets based on unique patient IDs. This resulted in 272 positive and 168 negative images in the training set from the GitHub dataset. To approximate an i.i.d. distribution in the training set, we added 2552 negative images from the NIH dataset, resulting in a total of 2992 training images (272 positive, 2720 negative). The test set consisted of 70 positive and 70 negative images, with 45 negative images from the NIH dataset and 25 from the GitHub dataset. This resulted in 2552 NIH images and 440 GitHub images in the training set, and 45 NIH images and 95 GitHub images in the test set.

\subsection{ML Models and Data for Anomaly Detection}

The \textit{D2Neighbors} model used in our experiments is an instance of the family of distance-based anomaly detectors, which encompasses a variety of methods from the literature \cite{HARMELING20061608,DBLP:books/sp/Aggarwal2013,DBLP:journals/csur/ChandolaBK09,DBLP:journals/pieee/RuffKVMSKDM21,DBLP:journals/csur/PangSCH21,DBLP:conf/icpr/RippelMM20}. The {D2Neighbors} model computes anomaly scores as
$o(\bx) = \mathbb{M}_j^{\gamma} \big\{\|\bx - \bsv_j\|_p^p\big\}$ where $\bx$ is the input, $(\bsv_j)_{j=1}^N$ are the training data, and $\mathbb{M}^\gamma$ is a generalized $f$-mean, with function $f(t) = \exp(-\gamma t)$. The function can be interpreted as a soft minimum over distances to data points, i.e.\ a distance to the nearest neighbors. In our experiments, the data received as input are RGB images of size $224 \times 224$ with pixel values encoded between $-1$ and $1$, downsized from their original high resolution using OpenCV's fast nearest neighbor interpolation. We set $\gamma$ so that the average perplexity \cite{10.1121/1.2016299} equals 25\% of the training set size for each model.

We also consider the {PatchCore} \cite{DBLP:conf/cvpr/RothPZSBG22} anomaly detection model, which uses mid-level patch features from a fixed pre-trained network. It constructs a memory bank of these features from nominal example images during training. Anomaly scores for test images are computed by finding the maximum distance between each test patch feature and its nearest neighbor in the memory bank. Distances are computed between patch features $\phi_p(\bx)$ and a memory bank of location independent prototypes $(\bsv_j)_{j=1}^N$. The overall outlier scoring function of PatchCore can be written as $o(\bx) = \max_{k}\min_{j} \|\phi_k(\bx) - \bsv_j\|$. The function $\phi_{k}$ is the feature representation aggregated from two consecutive layers at spatial patch location $k$, extracted from a pre-trained WideResNet50. The features from consecutive layers are aggregated by rescaling and concatenating the feature maps. The difference between our reported F1 scores and those in \cite{DBLP:conf/cvpr/RothPZSBG22} is mainly due to the method used to resize the images. We used the authors' reference implementation \footnote{\url{https://github.com/amazon-science/patchcore-inspection}} as the basis for our experiments.

All of the above models were trained on the MVTec dataset. The MVTec dataset consists of 15 image categories (`bottle', `cable', `capsule`, 'carpet', `grid', `hazelnut', `leather', `metal nut', `pill', `screw', `tile', `toothbrush', `transistor', `wood' and `zipper') of industrial objects and textures, with good and defective instances for each category. For the experiments based on D2Neighbors, we simulated different data preprocessing conditions before and after deployment by changing the way images are resized from their original high resolution to $224 \times 224$ pixels. We first use a resizing algorithm found in OpenCV 4.9.0 \cite{opencv_library} based on nearest neighbor interpolation. We then simulate post-deployment conditions using an improved resizing method, specifically a bilinear interpolation implemented in Pillow 10.3.0 and used by default in torchvision 0.17.2 \cite{torchvision2016}. This improved resizing method includes antialiasing, which has the effect of smoothing the transitions between adjacent pixels of the resized image.

\subsection{Explanations for Representation Learning}

Our experiments examined dot product similarities in representation space, i.e.\ $y = \langle \Phi(\bx), \Phi(\bx') \rangle$, where $\Phi$ denotes the function that maps the input features to the representation, typically a deep neural network. To understand similarity scores in terms of input features, we used the BiLRP method \cite{DBLP:journals/pami/EberleBKMVM22} which extends the LRP technique \cite{bach-plos15,DBLP:series/lncs/MontavonBLSM19,samek2019explainable,samek2021explaining} for this specific purpose. The conceptual starting point of BiLRP is the observation that a dot product is a bilinear function of its input. BiLRP then proceeds by reverse propagating the terms of the bilinear function to pairs of activations from the layer below and iterating down to the input. Denoting $R_{kk'}$ the contribution of neurons $k$ and $k'$ to the similarity score in some intermediate layer in the network, BiLRP extracts the contributions of pairs of neurons $j$ and $j'$ in the layer below via the propagation rule:
\begin{linenomath}
\begin{align}
R_{jj'} = \sum_{kk'} \frac{z_{jk}z_{j'k'}}{\sum_{jj'} z_{jk}z_{j'k'}} R_{kk'}
\end{align}
\end{linenomath}
In practice, this reverse propagation procedure can be implemented equivalently, but more efficiently and easily,  by computing a collection of standard LRP explanations (one for each neuron in the representation layer) and recombining them in a multiplicative manner:
\begin{linenomath}
\begin{align}
\text{BiLRP}(y) = \sum_{k} \text{LRP}(\Phi_k(\bx)) \otimes \text{LRP}(\Phi_k(\bx'))
\end{align}
\end{linenomath}
Overall, assuming the input consists of $d$ features, BiLRP produces an explanation of size $d \times d$ which is typically represented as a weighted bipartite graph between the set of features of the two input images. Due to the large number of terms, pixel-to-pixel contributions are aggregated into patch-to-patch contributions, and elements of the BiLRP explanations that are close to zero are omitted in the final explanation rendering. In our experiments, we computed BiLRP explanations using the Zennit\footnote{\url{https://github.com/chr5tphr/zennit}} implementation of LRP which handles the ResNet50 architecture, and set Zennit's LRP parameters to their default values.

\subsection{Explanations for the D2Neighbors Model}

The D2Neighbors model we investigate for anomaly detection is a composition of a distance layer and a soft min-pooling layer. To handle these layers, we use the purposely designed LRP rules of \cite{DBLP:journals/pr/KauffmannMM20,Montavon2022}. Propagation in the softmin layer ($\mathbb{M}_j^{\gamma}$) is given by the formula
\begin{linenomath}
\begin{align}
R_j &= \frac{f(\|\bx - \bsv_j\|_p^p)}{\sum_j f(\|\bx - \bsv_j\|_p^p)} o(\bx)
\end{align}
\end{linenomath}
a `min-take-most' redistribution, where $f$ is the same function as in $\mathbb{M}_j^{\gamma}$. Each score $R_j$ can be interpreted as the contribution of the training point $\bsv_j$ to the anomaly of $\bx$. To further propagate these scores into the pixel frequency domain, we adopt the framework of `virtual layers' \cite{vielhaben23,chormai24} and adapt it to the D2Neighbors model. As a frequency basis, we use the discrete cosine transform (DCT) \cite{10.3389/fnbot.2021.767299}, shown in  Fig.\ \ref{fig:vlayer} (right), which we denote by its collection of basis elements $(\bv_k)_k$. Since the DCT forms an orthogonal basis, we have the property $\sum_k \bv_k \bv_k^\top = I$, and multiplication by the identity matrix can be interpreted as a mapping to the frequencies and back. For the special case where $p=2$, the distance terms in D2Neighbors reduce to the squared Euclidean norm $\|\bx-\bsv_j\|^2$. These terms can be developed to identify pixel-pixel-frequency interactions: $\|\bx - \bsv_j\|^2 = (\bx - \bsv_j)^\top (\sum_k \bv_k \bv_k^\top) (\bx - \bsv_j) = \sum_k \sum_{ii'} [\bx - \bsv_j]_i [\bx - \bsv_j]_{i'} [\bv_k]_i [\bv_k]_{i'}$. From there, one can construct an LRP rule that propagates the instance-wise relevance $R_j$ to the pixel-frequency features:
\begin{linenomath}
\begin{align}
R_{ii'k} &= \sum_j \frac{[\bx - \bsv_j]_i [\bx - \bsv_j]_{i'} [\bv_k]_i [\bv_k]_{i'}}{\epsilon + \|\bx - \bsv_j\|^2} R_j,
\end{align}
\end{linenomath}
where the variable $\epsilon$ is a small positive term that handles the case where $\bx$ and $\bsv_j$ overlap. A reduction of this propagation rule can be obtained by marginalizing over interacting pixels ($R_{ik} = \sum_{i'} R_{ii'k}$). Further reductions can be obtained by marginalizing over pixels ($R_{k} = \sum_{i} R_{ik}$) or frequencies ($R_i = \sum_{k} R_{ik}$). These reductions are used to generate the heatmaps in Fig.\ \ref{fig:joint}.

\subsection{Data Availability}
All data used in this paper, in particular, NIH's CXR8 \cite{Wang2017}, GitHub-COVID \cite{cohen2020covidProspective}, ImageNet \cite{DBLP:conf/cvpr/DengDSLL009}, and MVTec-AD \cite{DBLP:conf/cvpr/BergmannFSS19} are publicly available.

\subsection{Code Availability}
Full code for reproduction of our results is available at \url{https://git.tu-berlin.de/jackmcrider/the-clever-hans-effect-in-unsupervised-learning}.

\subsection{Acknowledgements}

This work was partly funded by the German Ministry for Education and Research (under refs 01IS14013A-E, 01GQ1115, 01GQ0850, 01IS18056A, 01IS18025A and 01IS18037A), the German Research Foundation (DFG) as Math+: Berlin Mathematics Research Center (EXC 2046/1, project-ID: 390685689) and the DeSBi Research Unit (KI-FOR 5363, project ID: 459422098), and DFG KI-FOR 5363. Furthermore, KRM was partly supported by the Institute of Information \& Communications Technology Planning \& Evaluation (IITP) grants funded by the Korea government (MSIT) (No. 2019-0-00079, Artificial Intelligence Graduate School Program, Korea University and No. 2022-0-00984, Development of Artificial Intelligence Technology for Personalized Plug-and-Play Explanation and Verification of Explanation). We thank Stefan Ganscha for the valuable comments on the manuscript. Correspondence to KRM and GM. 

\renewcommand*{\bibfont}{\normalfont\small}
\printbibliography

\end{document}


\maketitle

\linenumbers

These supplementary notes provide detailed results and analysis of the experiments conducted in the main paper. They aim to further support the main paper's claims about the prominence and distinctiveness of the Clever Hans effect in unsupervised learning, and the importance of addressing it in order to achieve unsupervised models that are more robust and reliable for downstream applications.

\section{Additional Results for X-Ray Representations}

Recall from the main paper that we analyzed the PubMedCLIP's foundation model \cite{DBLP:conf/eacl/EslamiMM23}, specifically its representation of X-ray data and its use for a downstream COVID-19 detection task. For this purpose, we considered a dataset constructed by merging the NIH CXR8 dataset \cite{Wang2017} with the Github-hosted `COVID-19 Image Data Collection' \cite{cohen2020covidProspective}, and referred to these two data sources as NIH and Github. The NIH data, collected before the onset of the COVID-19 pandemic, is from a single hospital and is relatively homogeneous. The Github data, collected from multiple sources, is more heterogeneous. In the main paper, we reported strong inhomogeneities in the classification of the data across different subgroups, with the NIH instances being systematically correctly classified as negative and, in contrast, numerous Github negative instances being classified as positive. We attributed this heterogeneity in prediction quality to a spurious reliance on textual or annotation artifacts already latent in the representation of the unsupervised PubMedCLIP model. In this supplementary note, we retrace the emergence of the heterogeneous classification behavior from the representation level up to the classifier by providing further examples and analyses.

\subsection{Analysis of the Unsupervised Representation}

\begin{figure}[t!]
\centering
\includegraphics[width=\linewidth]{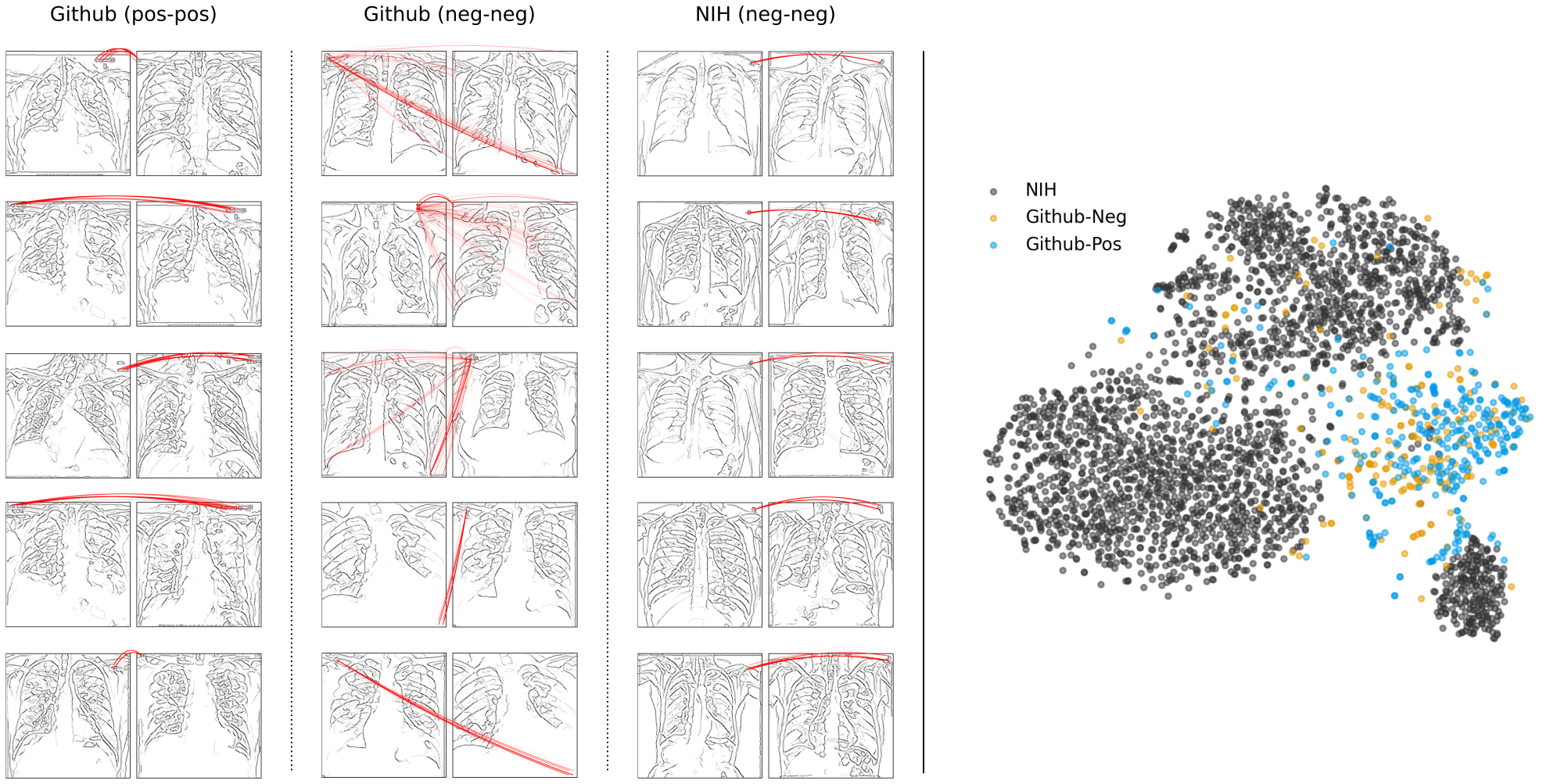}
\caption{Left: BiLRP explanations of PubMedCLIP's dot-product similarity scores for X-ray image pairs belonging to the same class and data subset. Red curves indicate pairs of features of the two images that support the similarity prediction. Right: T-SNE analysis of PubMedCLIP's representation of X-ray data. Colors indicate the data source (NIH CXR8 \cite{Wang2017} or Github-Covid \cite{cohen2020covidProspective}) and the ground truth label (COVID-positive or COVID-negative).}
    \label{fig:unsup-xray}
\end{figure}

In the main paper, we proposed to look at representation from the perspective of similarity between pairs of instances, and we expressed similarity as dot products in representation space (e.g.\ \cite{muller2001introduction}):
\begin{align}
k(x,x') = \Phi(x)^\top \Phi(x')
\label{eq:unsupervised}
\end{align}
The function $\Phi$ represents PubMedCLIP's mapping from raw pixel values to its representation. Fig.\ \ref{fig:unsup-xray} (left) shows the output of our BiLRP analysis, highlighting pairs of features responsible for the high measured dot-product similarity. Our analysis systematically highlights that the predicted similarity is supported by spurious annotations on the X-ray images instead of the underlying pathology, a clear case of a Clever Hans effect. In other words, the PubMedCLIP representation supports similarity predictions that are right for the wrong reasons. Furthermore, the type and location of these artifacts used by PubMedCLIP vary across the subgroups of the dataset, with the artifacts in the NIH subgroup being much more homogeneous than those in the Github subgroup.

The reliance of the PubMedCLIP model on features that code for dataset-specific annotation artifacts rather than the underlying pathology can also be demonstrated by performing a t-SNE analysis \cite{vanDerMaaten2008} of its representation. We color code each data point in the t-SNE visualization according to its provenance (NIH or Github). The visualization is shown in Fig.\ \ref{fig:unsup-xray} (right). It reveals a cluster structure in the data representation, where the clusters are more predictive of data provenance than COVID negativity/positivity. While the COVID-positives cluster at rather specific locations in the t-SNE space, they overlap significantly with the COVID-negatives from the Github dataset. Finally, it should be noted that while color-coded T-SNE analysis (and other correlational analyses for that matter) can establish the prominence of an already identified artifact or source of heterogeneity in the representation, our Explainable AI analysis does not require prior knowledge of the data artifact and can instead assist a human in identifying it.

\subsection{Analysis of the Downstream Classifier}

We now examine the transmission of the CH effect from the representation learning model to the downstream classifier. We recall from the main paper that we built a linear readout $f(x) = w^\top \Phi(x) + \theta$ on top of the PubMedCLIP representation. Specifically, we have trained a linear SVM classifier on the COVID detection task, where $f(x)<0$ represents COVID negatives and $f(x)>0$ represents COVID positives. Note that the same SVM classifier can also be formulated in terms of dot-product similarities:
\begin{align}
f(x) = \sum_{i=1}^N \alpha_i k(x,x_i) + \theta
\label{eq:supervised}
\end{align}
The predictions of the SVM can thus be seen as a combination of dot-product similarities of the type we analyzed above. After training the SVM model and selecting the hyperparameters using hold-out validation, we found that $C=0.01$ and a class weighting scheme adjusted to balance the class representation yielded the best performance. We examine the SVM prediction performance over the different data subgroups. The results are shown in Table \ref{table:covid}. We observe a strong inhomogeneity of performance between different data sources.
\begin{table}[b!]
\small \centering
\renewcommand{\arraystretch}{1.15}
\begin{tabular}{lccc}
& original & github & NIH\\\midrule
Accuracy & 88.6   & 75.2 & 100.0\\
False positive rate (FPR) & 14\%  & 40\% &  0\% \\
 \end{tabular}
\caption{Analysis of a linear SVM built on the PubMedCLIP's unsupervised foundation model and trained on the COVID classification task. Accuracy and false positive rates are reported for the original data (aggregation of github and NIH sources), and for the individual data sources.}
\label{table:covid}
\end{table}

\begin{figure}[t!] \centering
    \includegraphics{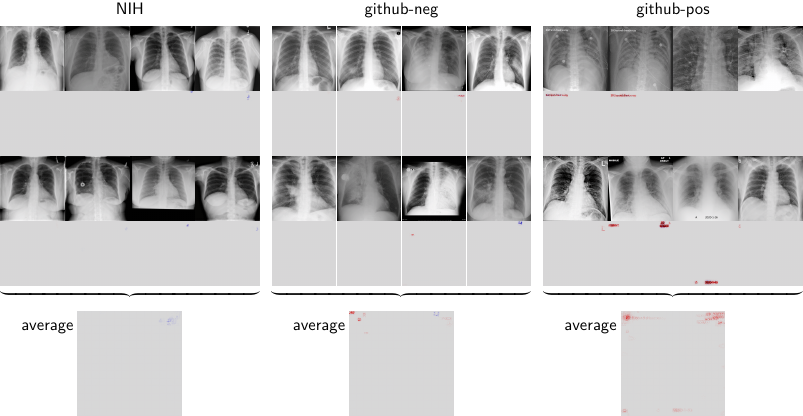}
    \caption{LRP explanations for COVID-19 detection. Top: Exemplary chest X-ray images from each data subgroup along with explanations showing pixel-wise evidence for/against COVID-19 as perceived by the model. Bottom: Explanations averaged over the entire subgroup. Red color indicates regions contributing evidence for COVID-19, while blue color indicates evidence against COVID-19.}
    \label{fig:lrp-xray}
\end{figure}

To further shed light on the instability in the SVM's decision strategy, we use the LRP explanation technique \cite{bach-plos15,DBLP:series/lncs/MontavonBLSM19}. LRP allows us to identify, for each predicted instance, the extent to which each input pixel contributed to the prediction. The results are shown in Fig.\ \ref{fig:lrp-xray} for a selection of correctly classified instances from each data subgroup.
Similar to the BiLRP analysis above, our LRP analysis of the SVM classifier reveals a strong reliance of the model on spurious annotations, with COVID-positive instances again being the ones most strongly supported by artifacts. The LRP analysis also reveals that COVID-negative predictions are supported by annotation artifacts. The latter, however, differ from the positive ones in that they systematically occur in the upper right corner of the input image. This Clever Hans-like prediction strategy, combined with the fact that the artifacts differ across datasets, explains the unstable prediction performance of the classifier. In particular, the high false positive rate (FPR) of the SVM on the Github data can be explained by the fact that the artifacts of the Github negatives do not coincide with the artifacts of the NIH negatives, with the latter constituting the majority of the negative instances. As a result, Github negatives fail to be drawn to the negative class and often end up being misclassified as positives. In summary, our analysis of the downstream classifier further verifies and scrutinizes the inheritance of the CH effect from the unsupervised model to the supervised model and its consequences in terms of prediction performance.

\section{Additional Results for Generic Image Models}

In the main paper, we also experimented with generic image models trained with different unsupervised representation learning algorithms, specifically \textit{SimCLR} \cite{simclr-v1, simclr-v2}, \text{BarlowTwins} \cite{DBLP:conf/icml/ZbontarJMLD21} and \textit{CLIP} \cite{DBLP:conf/icml/RadfordKHRGASAM21}, the details of which can be found in the methods section of the main paper. We extend on these experiments by providing further analysis of the learned representations and the downstream classifiers built on top of them.

\subsection{Analysis of the Unsupervised Representations}

\begin{figure}[b!]
    \makebox[\textwidth][c]{
    \includegraphics[width=\linewidth]{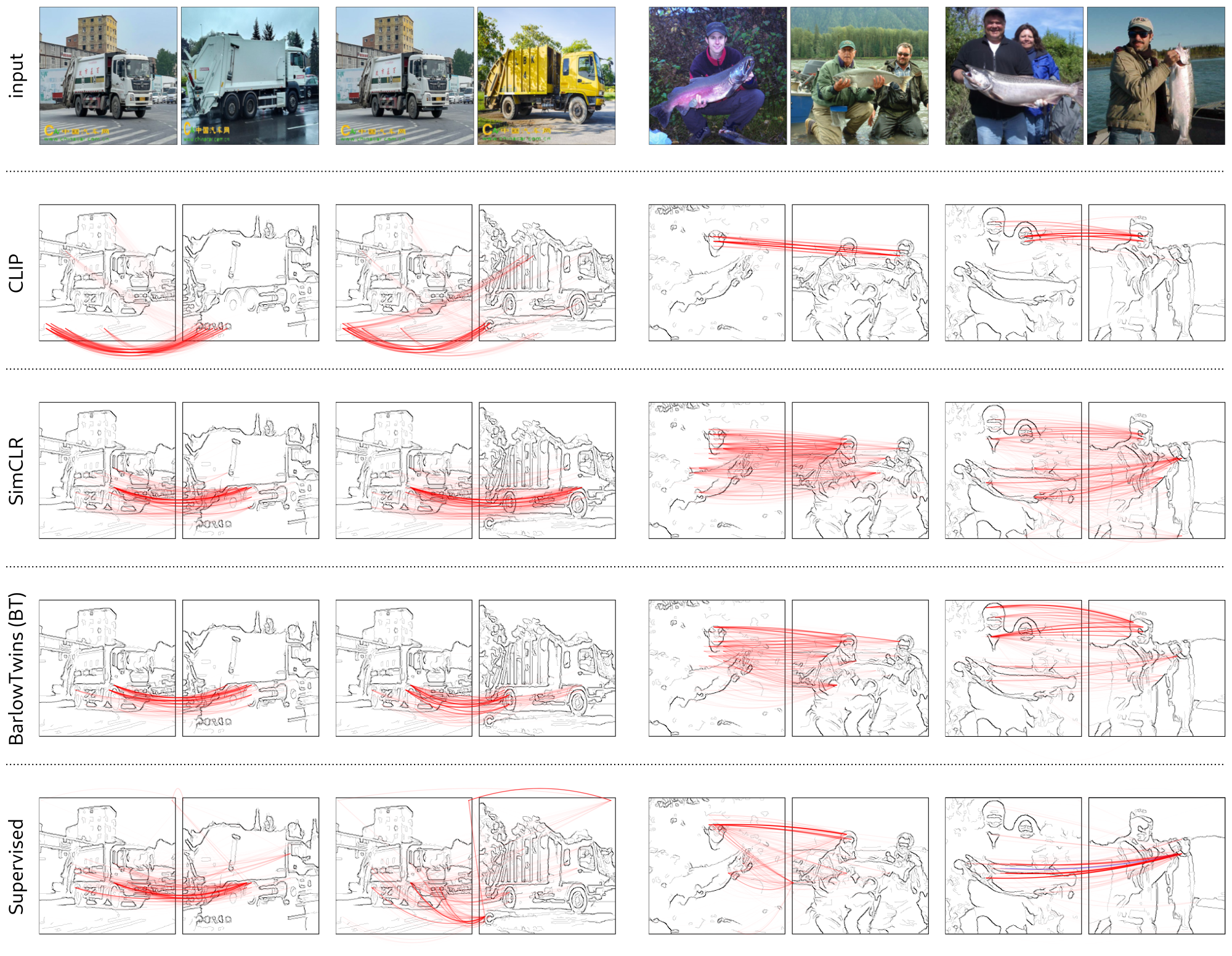}
    }
    \caption{BiLRP explanations of dot product similarities for each representation learning model. The explanations are computed on a selection of ImageNet images from the truck and fish subsets. We observe several Clever Hans effects, such as a reliance on copyright tags in the truck subset and a reliance on humans in the fish subset.}
    \label{fig:bilrp-fish-truck}
\end{figure}

We first proceed with the BiLRP analysis of each model as well as a purely supervised model (a standard ResNet50 pre-trained on ImageNet). The results of the analysis are shown in Fig.\ \ref{fig:bilrp-fish-truck} for additional image pairs of the class garbage truck and coho, respectively.

Recall our findings from the main paper that similarity predictions built on unsupervised representations are often Clever Hans-like. For example, the similarity between image pairs of certain fish classes was supported by a human holding the fish in the background (either the human face for the CLIP model, or the human body for SimCLR and BarlowTwins). Similarly, for the truck data, the CLIP model was shown to rely heavily on textual logos to express similarity. Highlighting the heterogeneity of the unsupervised representations, the same textual logos were suppressed by SimCLR and BarlowTwins, presumably due to their eccentricity in the image. Here, we complement these results by including a supervised baseline in our analysis (leftmost example in Fig.\ \ref{fig:bilrp-fish-truck}). We find that the supervised baseline model focuses on task-relevant features better than unsupervised approaches. However, the supervised baseline also appears to be overfitted to the task at hand, matching features that belong to the same class, but failing to disentangle distinct object parts, such as the wheels and the roof of the truck.

\begin{figure}[t!]
    \includegraphics[width=\textwidth]{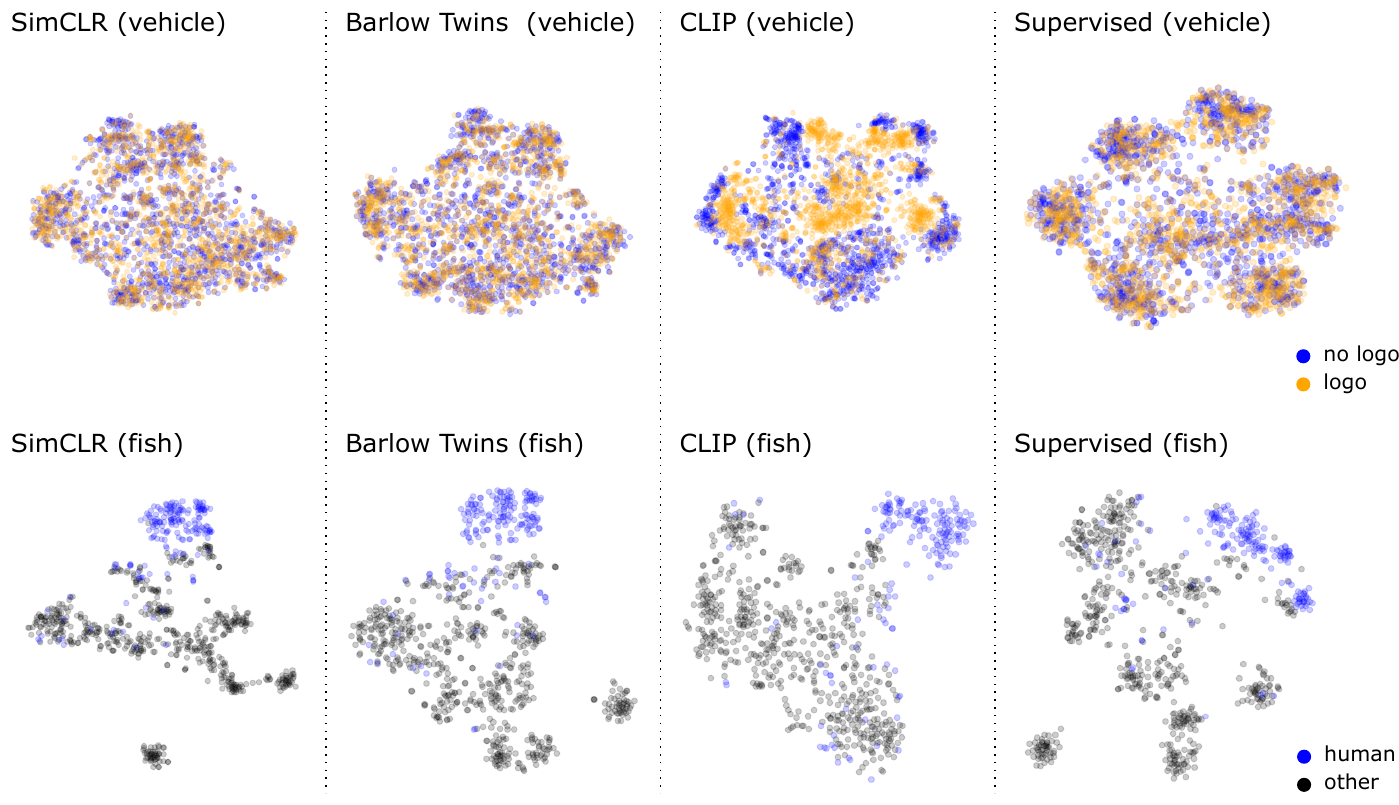}
    \caption{T-SNE plots of the representations from different models of the truck subset (top) and the fish subset (bottom). For the truck subset, instances are colored according to the presence or absence of a textual logo in the image. For the fish subset, instances are colored according to the presence of a human in the image.}
    \label{fig:tsne-fish-truck}
\end{figure}

To further verify the result of the BiLRP analysis, which among other things showed that textual logos are salient in the CLIP representation, we perform a t-SNE analysis \cite{vanDerMaaten2008} in representation space. We append to the original truck data an artificial dataset called `logo', consisting of truck data with a textual logo artificially inserted into each image. We then compute a t-SNE embedding on the augmented data, mapped to the representation of each model. Figure \ref{fig:tsne-fish-truck} shows the t-SNE plots for each model's representation, color-coded by the presence or absence of a logo. For the CLIP model, the logo is strongly expressed in its representation, as shown by the yellow and blue dots forming distinct clusters. This suggests that a downstream classifier has an incentive to use the textual logo as part of its prediction strategy. However, for the SimCLR and Barlow Twins models, as well as for the supervised baseline, the analysis suggests that the logo does not play a significant role.

We repeat the same t-SNE analysis for the `fish' subset, with the resulting embedding color-coded according to the presence or absence of a human. We use a Faster R-CNN object detection model pre-trained on the COCO dataset \cite{ren2015faster} to classify each image according to whether it contains a human or not. In each case, we observe high representational similarity of images containing humans. This suggests that human features are easily accessible when performing the downstream fish classification task.

\subsection{Analysis of Downstream Classifiers}

\begin{figure}[t!]
\centering
\includegraphics[width=\textwidth]{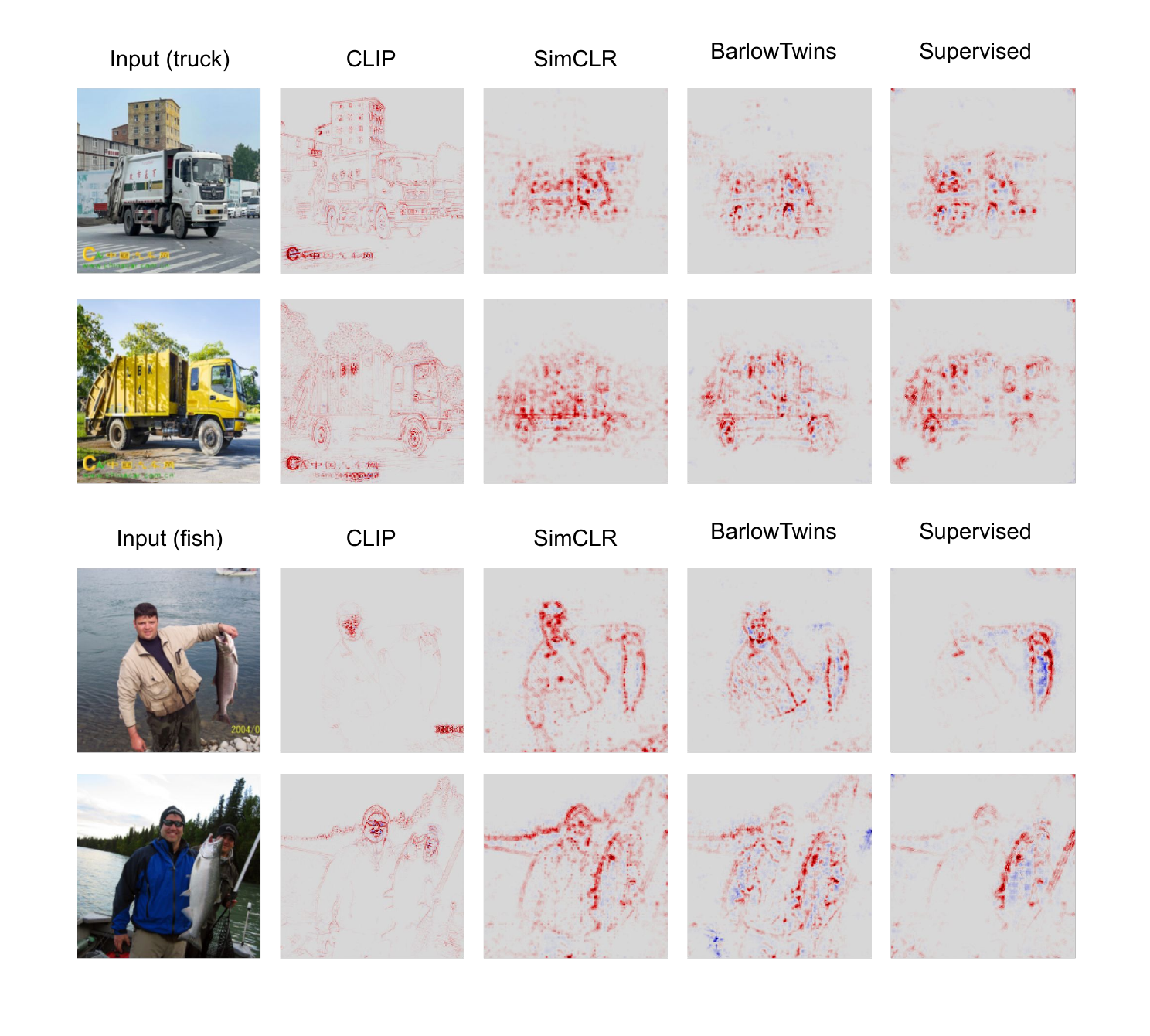}
	\caption{LRP heatmaps for the downstream classification of the fish and truck subsets for different representation learning models. The heatmaps show the pixel-wise contributions to the output of the downstream classifier for the ground truth class. Red color indicates positively contributing pixels and blue color indicates negatively contributing pixels.}
	\label{fig:lrp-imagenet}
\end{figure}

\begin{figure}[b!]
\centering
\includegraphics[width=\textwidth]{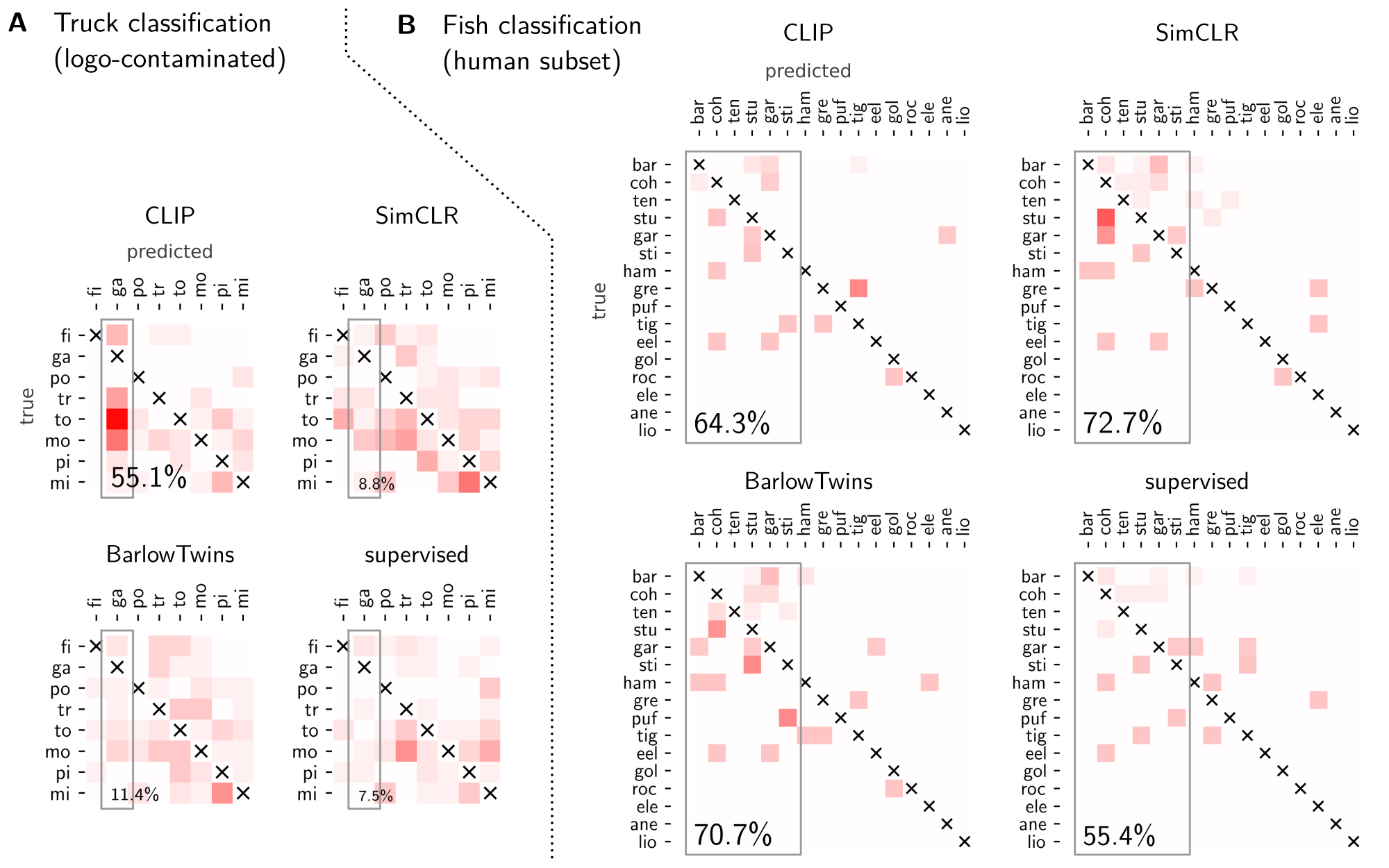}
\caption{Confusion matrices highlighting the distribution of model errors in terms of true and predicted class. The percentages shown correspond to the proportion of errors falling in the indicated region of the confusion matrix. Abbreviations:
[fi]re\,engine, [ga]rbage\,truck, [po]lice\,van, [tr]ailer\,truck, [to]w\,truck, [mo]ving\,van, [pi]ckup, [mi]nivan
for the truck classes, and
[bar]racouta, [coh]o, [ten]ch, [stu]rgeon, [gar], [sti]ngray, [ham]merhead, [gre]at\,white\,shark, [puf]fer, [tig]er\,shark, [eel], [gol]dfish, [roc]k\,beauty, [ele]ctric\,ray, [ane]mone\,fish, [lio]nfish
for the fish classes.
}
\label{fig:confusion-imagenet}
\end{figure}

To analyze the transmission of the unsupervised Clever Hans effect from the unsupervised representation to the downstream supervised models, in the main paper we investigated the construction of two supervised models on top of each representation (one for fish classification and one for truck classification). These models are linear-softmax classifiers trained with cross-entropy loss. We extend the analysis of these classifiers by analyzing their decision strategy using LRP heatmaps.  We use the same LRP procedure that we have used to compute BiLRP explanations, and we apply the rule LRP-0 \cite{DBLP:series/lncs/MontavonBLSM19} in the last readout layer. LRP heatmaps are shown in Fig.\ \ref{fig:lrp-imagenet}. We also report the accuracy scores of each model for each data subset considered in Table \ref{table:imagenet}. We further elaborate on the structure of the accuracy scores of the problematic data subgroups (truck images with a logo and fish images with a human) by providing the full confusion matrices in Fig.\ \ref{fig:confusion-imagenet}.

We start with the truck classification task, where we distinguish between the original data, consisting of the 50 test images of each truck (400 images in total), and the logo subset, derived from the original test data by artificially inserting in each image a logo similar to the one found in the `garbage truck' class. On the original data, we observe that the CLIP model performs best with an accuracy of 86.6\%, followed by 85.7\% for the supervised baseline, and sensibly lower accuracies of 82.9\% and 77.7\% for the BarlowTwins and SimCLR models, respectively (cf.\ Table \ref{table:imagenet}, column `original'). However, the high accuracy of the CLIP model is partly the result of a CH strategy based on the detection of a logo in the lower left corner of the image, as shown by the LRP heatmaps in Fig.\ \ref{fig:lrp-imagenet}. In comparison, models built on the BarlowTwins and SimCLR representations, as well as our supervised baseline, are largely unresponsive to this logo. Notably, the reliance (or lack thereof) of the supervised model mirrors our BiLRP analysis (cf.\ Fig.\ \ref{fig:bilrp-fish-truck}), where only the CLIP model was affected by the logo artifact.

Looking at the prediction accuracy on the subgroup of truck images with artificially added logos (see Table \ref{table:imagenet}, column `logo'), the performance of SimCLR, BarlowTwins and our supervised baseline remains stable compared to the original test data, with a maximum performance drop of less than 1\%. On the other hand, CLIP's performance drops drastically by almost five percentage points, making it the worst performer on this subset. The reason for the low accuracy of the CLIP model is revealed by looking at CLIP's confusion matrix (cf.\ Fig.\ \ref{fig:confusion-imagenet}). One observes that a whopping 55.1\% of the CLIP model's errors are misclassifications into the `garbage truck' class, compared to less than 15\% for non-CLIP models.

\begin{table}[t!]
\small \centering
\newcommand{\pdecrease}[1]{$\phantom{\downarrow}$\,#1\,$\color{red}\downarrow$}
\renewcommand{\arraystretch}{1.15}
\begin{tabular}{lcc|ccc}
\multicolumn{1}{c}{}
& \multicolumn{2}{c}{ImageNet:truck} & \multicolumn{3}{c}{ImageNet:fish}\\\midrule
\multicolumn{1}{c}{} & \!\!\textit{original}\!\! & \multicolumn{1}{c}{logo} & \!\!\textit{original}\!\! & human & no human\\\midrule
CLIP &  \textit{85.0}    & \pdecrease{80.5} & \textit{86.5}       & 83.8    & 82.5\\
\rowcolor{gray!20!white}~+~CH mitigation (1) &  \textit{85.0} & 83.8 \\
\rowcolor{gray!20!white}~+~CH mitigation (2) &  \textit{84.8} & 84.0 \\
\rowcolor{gray!20!white}~+~CH mitigation (5) &  \textit{85.2} & 85.0 \\
\rowcolor{gray!20!white}~+~CH mitigation (10) &  \textit{85.0} & 85.0\\
\rowcolor{gray!20!white}~+~CH mitigation (20) &  \textit{81.2} & 81.2\\
 & & \\[-3mm]
SimCLR  &  \textit{74.8}  & 74.5  & \textit{82.2}   & 78.6 & 78.4  \\
BarlowTwins &  \textit{80.2}  & 80.2  & \textit{83.1}    & \pdecrease{75.6}  & 81.1 \\
\textit{Supervised}   &  \textit{83.8}  &  83.2     & \textit{86.2}  &  84.2 & 81.9  \\\bottomrule
\end{tabular}

\caption{Prediction accuracy obtained by downstream classification models built on the truck and fish subtasks from each representation (including representations enhanced for CH mitigation by removing a given number of spuriously relevant feature maps). Accuracy is reported for the original data, for truck images with an artificially added logo, and for the subgroups of fish images with and without humans.}
\label{table:imagenet}
\end{table}

We now investigate downstream models that classify the fish ImageNet subset from the different unsupervised representations. All models perform reasonably well on this classification task. The CLIP model ranks first with 86.5\%, followed by the supervised model (86.2\%), and BarlowTwins and SimCLR with 83.1\% and 82.2\%, respectively. Figure \ref{fig:lrp-imagenet} shows LRP heatmaps explaining the classification by each model of a selection of images from the class `coho'. We observe that the CLIP model focuses on the human in the image, specifically the face, rather than the fish. Similarly, the SimCLR and BarlowTwins models also focus on the human, but more specifically on the body. Only the supervised model is reasonably unresponsive to the human and instead clearly focuses on the actual fish to be classified. This shows that the unsupervised learning models studied here all implement a CH strategy. This casts doubt on the generalization of these models and their performance across different subgroups.

We investigate the consequence of the identified CH strategies by building a specific subgroup consisting of all fish instances containing humans (e.g.\ images with a human holding a fish in his hands). This subgroup is built by including only images for which the Faster-R CNN auxiliary model has detected the presence of a human. Since humans are not homogeneously distributed across classes in both training and test data, we rebalance the test data by reweighting all classes containing more than 10 instances in a way that their effective sample size becomes 10. This rebalancing can also be interpreted as a simulation of real-world conditions, removing potentially spurious correlations between fish classes and the presence of a human. On this class-rebalanced subset of data, the accuracy of SimCLR drops below 80\%, and BarlowTwins achieves an accuracy of only 75.6\%.

The decrease in performance for fish classification, as for the truck case, can be understood by looking at the confusion matrices (see Fig.\ \ref{fig:confusion-imagenet}). We observe that more than 70\% of the errors of the SimCLR and BarlowTwins models correspond to misclassifications as one of the six classes with the highest human prevalence in the original data (`barracouta', `coho', `tench', `sturgeon', `gar', `stingray'). In contrast, only 55.4\% of the errors of the supervised model correspond to this type of misclassification. The reason for the distinct error structure of the SimCLR and BarlowTwins models is that their weak representation of fish features, as shown in the LRP and BiLRP explanations, induces the model to exploit the spurious correlation between human presence and these six classes. As a result, any fish image that contains a human will tend to be predicted, often incorrectly, into one of these classes.

\section{Additional Results for Anomaly Detection}

This note complements the unsupervised anomaly detection experiments from the main paper. It provides the Explainable AI analysis and measured performance before and after CH mitigation, for each MVTec category retained in our experiments. Note that our criterion for retaining an MVTec category was that the D2Neighbors model accurately predicts it with F1 scores above 0.9 on the original data (cf.\ Table \ref{table:anomaly}, column \textit{original}).

\begin{figure*}[t!]
    \centering
    \includegraphics[page=2]{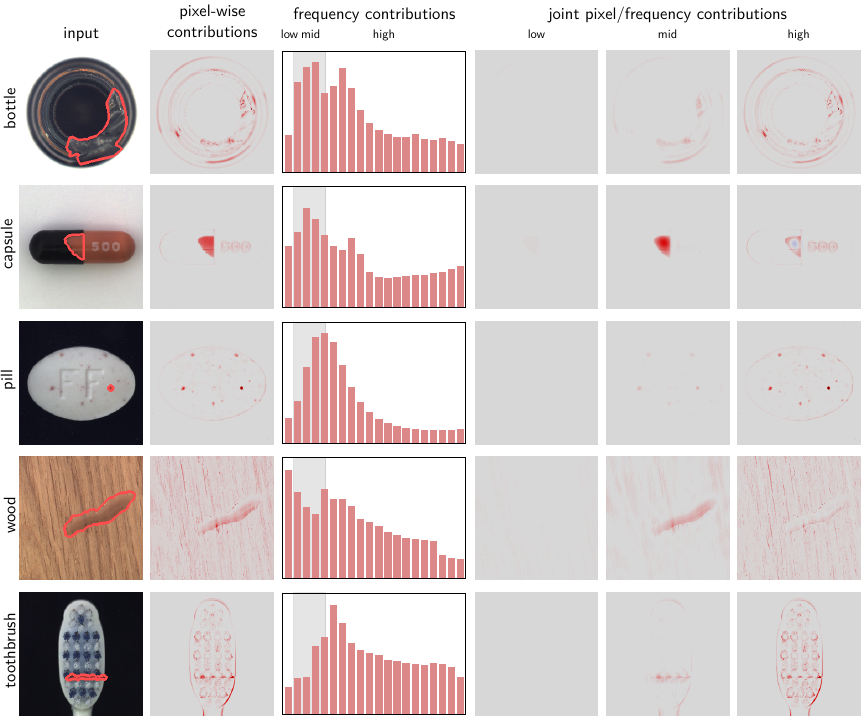}
    \caption{For the retained 5 categories of the MVTec-AD dataset, we show  (from left to right), anomalous input images (with ground truth outlined in red), pixel-wise explanations of the D2Neighbors anomaly predictions, frequency-based explanations, and joint pixel-frequency explanations. For the frequency-based explanations, the x-axis is plotted on a power scale, with bins representing increasingly larger frequency ranges as we move to the high frequencies: (0-2), (3-17), (18-69), (70-188), (189-409), (410-769), (770-1313), (1314-2088), (2089-3142), (3143-4528), (4529-6303), (6304-8525), (8526-11254), (11255-18491), (18492-23132), (23133-28548), (28549-34811), (34812-41995), (41996-50176).}
    \label{fig:anomaly_overview}
\end{figure*}

To verify the nature of these accurate anomaly predictions, and in particular to test whether they are based on a valid strategy or instead `right for the wrong reasons', we applied an Explainable AI analysis in the main paper. Here we provide the details of the analysis for the five retained MvTec categories and their corresponding D2Neighbors model. The analysis provides pixel-wise explanations of the predicted anomaly scores. By introducing a Discrete Cosine Transform (DCT) virtual layer, as described in the main paper, the analysis is extended to produce frequency and joint pixel-frequency explanations. Explanations for typical instances of each retained MVTec-AD category are shown in Figure \ref{fig:anomaly_overview}.

For each MvTec category, the D2Neighbors model responds to a wide range of frequencies, with different frequency bands contributing differently to the decision strategy. In particular, the true anomaly patterns are mostly expressed in the mid-frequency range, while the low and high frequency bands contain artifacts. Especially in the high frequency range, we can see the model's response to the border of the bottle, text markings on the capsule, specks on the pill, wood stripes, or the individual hairs of the toothbrush, all of which are typically not indicative of a manufacturing defect (the true manufacturing defect is marked with a red outline in the input image). In other words, our analysis shows that part of D2Neighbors' anomaly prediction strategy is of the Clever Hans type.

\begin{figure}[t!]
\centering
\includegraphics{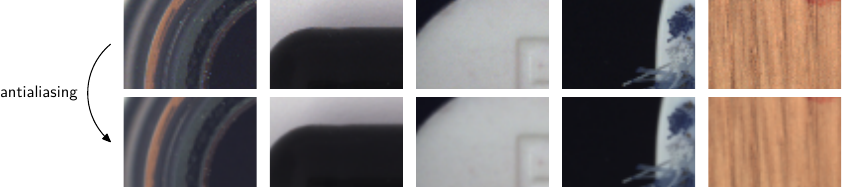}
\caption{Effect of the resizing procedure on a selection of images from each retained MVTec category. Top: original resizing procedure without antialiasing. Bottom: simulated post-deployment resizing procedure with antialiasing. We can observe that the post-deployment images are sensibly less noisy.}
\label{fig:mvtec-images}
\end{figure}

In the main paper, we investigated the consequences of the identified erroneous decision strategy on the prediction performance under a change in data quality. Specifically, the D2Neighbors model was trained and validated on MVTec images resized using a coarse nearest neighbor interpolation algorithm, and we simulated a post-deployment scenario in which a better bilinear resizing incorporating antialiasing is being used (details in the methods section of the main paper). The effect of changing the resizing procedure for one representative image of each MVTec category is shown in Figure \ref{fig:mvtec-images}.

We observe that the introduction of antialiasing in the resizing procedure strips images of their high-frequency noise-like components. This suggests that the output of an anomaly model that is sensitive to high frequencies may be significantly affected, especially if the model's exposure to high frequencies is not confined to a particular region of the input image. This also suggests that the performance of a downstream model, such as a detector of manufacturing flaws, may be severely degraded. This is confirmed in Table \ref{table:anomaly} where we report per category and averaged F1 scores before and after modifying the resizing algorithm (columns `original' and `deployed'). In particular, we observe that D2Neighbors' F1 score drops significantly after the resizing change, and this effect is generally consistent across categories. In particular, we observe a severe performance degradation on the `wood' category, which can be explained by the spatially widespread presence of high-frequency artifacts in the original conditions (cf.\ Fig.\ \ref{fig:mvtec-images}).

\begin{table}[t!]
\centering
\newcommand{\pdecrease}[1]{$\phantom{\downarrow}$\,#1\,$\color{red}\downarrow$}
\newcommand{\pincrease}[1]{$\phantom{\uparrow}$\,#1\,$\color{green!50!black}\uparrow$}
\renewcommand{\arraystretch}{1.15}

\begin{tabular}{lcccc}
& \textit{original} & \scriptsize (bot, cap, pil, too, woo) & deployed &  \scriptsize (bot, cap, pil, too, woo)\\\midrule
D2Neighbors ($\ell_1$) & \textit{0.93} & \scriptsize (0.89, 0.91, 0.94, 0.97, 0.93) & \pdecrease{0.74} & \scriptsize (0.87, 0.45, 0.83, 0.87, 0.71) \\
\rowcolor{gray!20!white}~+~CH mitigation & \textit{0.92} & \scriptsize (0,88, 0.91, 0.93, 0.97, 0.94) & \pincrease{0.92} & \scriptsize (0.88, 0.91, 0.93, 0.97, 0.94) \\
\\[-3mm]
D2Neighbors ($\ell_2$) & \textit{0.92} & \scriptsize (0.90, 0.91, 0.92, 0.93, 0.93) & \pdecrease{0.80} & \scriptsize (0.85, 0.73, 0.91, 0.85, 0.67)\\
\rowcolor{gray!20!white}~+~CH mitigation & \textit{0.94} & \scriptsize (0.91, 0.92, 0.92, 0.98, 0.95) & \pincrease{0.93} & \scriptsize (0.91, 0.92, 0.92, 0.98, 0.93) \\
\\[-3mm]
D2Neighbors ($\ell_4$) & \textit{0.91} & \scriptsize (0.92, 0.90, 0.92, 0.90, 0.93) & \pdecrease{0.83} & \scriptsize (0.83, 0.88, 0.89, 0.75, 0.78) \\
\rowcolor{gray!20!white}~+~CH mitigation & \textit{0.93} & \scriptsize (0.93, 0.90, 0.91, 0.98, 0.93) & \pincrease{0.93} & \scriptsize (0.93, 0.90, 0.91, 0.98, 0.92) \\
\\[-3mm]
PatchCore &\textit{0.92} & \scriptsize (0.92, 0.90, 0.92, 0.89, 0.98) & \pdecrease{0.85} & \scriptsize (0,91, 0.88, 0.88, 0.64, 0.96) \\
\rowcolor{gray!20!white}~+~CH mitigation & \pincrease{\textit{0.97}} & \scriptsize (1.00, 0.96, 0.93, 1.00, 0.95) & \pincrease{0.96} & \scriptsize (1.00, 0.95, 0.93, 0.98, 0.95) \\
\bottomrule
\end{tabular}
\caption{F1 scores on the MVTec-AD data per category and averaged. Scores are reported for the simulated original and post-deployment conditions and for a variety of anomaly detection models. Abbreviations: [bot]tle, [cap]sule, [pil]l, [too]thbrush, [woo]d.}
\label{table:anomaly}
\end{table}

Furthermore, the effect of antialiasing varies between different norm configurations in the D2Neighbors models. The $\ell_1$-norm variant, which has a higher overall exposure than the $\ell_2$-norm variant, sees its F1 score drop from 0.93 to 0.74 after antialiasing is introduced. Again, the wood category is severely affected, but even more so the capsule category, where the fine asperities of the surface on which the capsule is placed cause high frequency noise all over the image. Note, however, that since the anomalies for the capsule category are more spatially confined than those for the wood category, the $\ell_2$ and $\ell_4$-norm variants of D2Neighbors perform comparatively much better after deployment due to their pixel-sparsity mechanism. The $\ell_4$-norm variant generally shows the strongest resilience across all categories, with an average F1 score that drops from 0.91 to a less severe 0.83.

The difficulty of the D2Neighbors model in providing anomaly scores that generalize to changing data conditions also occurs in PatchCore, where we observe a decrease in the F1 score from 0.92 on the original data to 0.85 after deployment. This is significant because D2Neighbors and PatchCore are very different in structure. The latter extracts features using a pre-trained WideResNet50 before computing distances, and applies spatial max-pooling to identify the region in the image that is most responsible for the anomaly, building invariance to the other regions. The effect of these structural differences can be seen by looking at the F1 scores per category. In contrast to D2Neighbors, PatchCore experiences the largest decrease in performance for the `toothbrush' class (from 0.89 to 0.64). A possible explanation is an amplification/suppression effect in the PatchCore feature extractor, similar to those found in CLIP, SimCLR, and BarlowTwins. Specifically, PatchCore reacts strongly to hair pattern of the toothbrush, and the very thin anomaly-causing bent hairs no longer find their way through PatchCore's feature extractor after being weakened by antialiasing.

The Clever Hans effect, where models exploit spurious correlations rather than robust patterns, manifests as overfitting to high-frequency noise in training data, leading to poor generalization on anti-aliased test images.  In the main paper, we demonstrated the significant performance degradation of CH models under changing data conditions and provided detailed results in Table \ref{table:anomaly}. We then proposed a simple but effective CH mitigation strategy consisting of inserting a pre-processing layer at the input of each anomaly detection model that performs an 11x11 Gaussian blur. Note that the architecture modification is made before training, so that the data is systematically blurred during training and testing and also after deployment. The blurring layer is designed to reduce the model's response to high frequency noise while maintaining its response to low- and mid-frequency patterns. The effect in terms of F1 score of CH mitigation is shown in Table \ref{table:anomaly}.

We observe an improvement in performance on the deployed test set across all models and categories. For example, D2Neighbors $\ell_2$ recovers from 0.80 to 0.93 with CH mitigation. The per-category F1 scores show that the performance recovery is systematic across all categories, and even outperforms the original model for the `toothbrush' category. This last result suggests that blurring also improves D2Neighbors' ability to represent anomalies. Applying the same blur filter to the input of PatchCore, we observe that the F1 score jumps from 0.85 to 0.96. Again, this is a significant improvement over the original performance of 0.92. One hypothesis is that the blurred images propagate better through PatchCore's feature extraction layers, allowing a finer anomaly detection model to be built than the one built on noisy images. Overall, these results highlight the effectiveness of CH mitigation in improving both average and per-category accuracy for all models tested.

\renewcommand*{\bibfont}{\normalfont\small}
\printbibliography